\documentclass[conference]{IEEEtran}
\IEEEoverridecommandlockouts
\usepackage{cite}
\usepackage{amsmath,amssymb,amsfonts}
\usepackage{algorithmic}
\usepackage{graphicx}
\usepackage{textcomp}
\usepackage{xcolor}
\usepackage{hyperref}
\usepackage{booktabs}
\usepackage{multirow}
\def\BibTeX{{\rm B\kern-.05em{\sc i\kern-.025em b}\kern-.08em
    T\kern-.1667em\lower.7ex\hbox{E}\kern-.125emX}}
\begin{document}

\title{On the Adversarial Robustness of Graph Neural
Networks with Graph Reduction}

\author{\IEEEauthorblockN{Kerui Wu}
\IEEEauthorblockA{
\textit{Rensselaer Polytechnic Institute}\\
Troy, NY, USA\\
wuk9@rpi.edi}
\and
\IEEEauthorblockN{Ka-Ho Chow}
\IEEEauthorblockA{
\textit{University of Hong Kong}\\
Hong Kong, China\\
kachow@cs.hku.hk}
\and
\IEEEauthorblockN{Wenqi Wei}
\IEEEauthorblockA{
\textit{Fordham University}\\
New York, NY, USA\\
wenqiwei@fordham.edu}
\and
\IEEEauthorblockN{Lei Yu}
\IEEEauthorblockA{
\textit{Rensselaer Polytechnic Institute}\\
Troy, NY, USA\\
yul9@rpi.edu}
}

\maketitle

\begin{abstract}
As Graph Neural Networks (GNNs) become increasingly popular for learning from large-scale graph data across various domains, their susceptibility to adversarial attacks when using graph reduction techniques for scalability remains underexplored. In this paper, we present an extensive empirical study to investigate the impact of graph reduction techniques—specifically graph coarsening and sparsification—on the robustness of GNNs against adversarial attacks. Through extensive experiments involving multiple datasets and GNN architectures, we examine the effects of four sparsification and six coarsening methods on the poisoning attacks. 
Our results indicate that, while graph sparsification can mitigate the effectiveness of certain poisoning attacks, such as Mettack, it has limited impact on others, like PGD. Conversely, graph coarsening tends to amplify the adversarial impact, significantly reducing classification accuracy as the reduction ratio decreases. Additionally, we provide a novel analysis of the causes driving these effects and examine how defensive GNN models perform under graph reduction, offering practical insights for designing robust GNNs within graph acceleration systems.
\end{abstract}


\section{Introduction}
Today, graphs have grown exponentially in size and complexity, serving as fundamental and powerful data structures that depict a vast array of entities and their interconnections. From social networks and financial systems to transportation networks, the ability to effectively represent and analyze such large-scale graphs is crucial.
Graph neural networks (GNNs)\cite{hamilton2017inductive,kipf2016semi,velivckovic2017graph} have recently emerged as a pivotal technique for learning from and making predictions on graph data, with broad applications across various domains. Despite their growing popularity and wide application in diverse fields such as fraud detection~\cite{liu2021pick}, drug discovery\cite{sun2020graph}, and software vulnerability detection~\cite{mirsky2023vulchecker}, the scalability of GNNs has become a significant challenge due to the inherent requirement of GNNs on leveraging information from multi-hop neighbors to generate meaningful node representations and accurate predictions. To address the scalability issue of GNNs, multiple acceleration techniques, including graph coarsening, graph sparsification, and customized hardware design, have been applied to improve the efficiency of GNN learning on large graph data~\cite{zhang2023survey}.

On the other hand, numerous studies\cite{zugner2020nettack,mettack,dai2023unnoticeable} have shown that GNNs are vulnerable to adversarial attacks. By modifying the original graph by adding, removing edges, or perturbing node attributes, adversaries can dramatically mislead the classifiers and damage the model quality. To address such vulnerabilities, various methods, including preprocessing techniques~\cite{wu2019jaccard,entezari2020gcnsvd} and defensive models \cite{zhang2020gnnguard,zhu2019robustgcn,chen2021median}, have been proposed to improve the adversarial robustness of GNN. However, these attacks and defense methods are only tested on non-accelerated systems using vast original graphs. Although graph reduction methods such as coarsening and sparsification show promise as a means to accelerate training processes, their impact on GNNs' robustness has not been thoroughly investigated. Previous work~\cite{zhu2024robustness} studied the effect of graph reduction on GNN robustness but focused only on graph backdoor attacks. The impact of graph reduction on more general adversarial attacks and evasion attacks is still unexamined. Addressing this gap is crucial for building more robust GNN systems that are both resistant to adversarial attacks and efficient in handling large-scale graphs.

In this paper, we systematically examine the impact of graph reduction on the robustness of GNNs against various adversarial attacks\cite{mettack,xu2019pgd,geisler2021prbcd,liu2023grad,li2022strg,bojchevski2019nea} using diverse attack strategies. Specifically, we aim to address three critical questions: 
(1) How do poisoning attacks affect GNN performance in accelerated systems employing graph reduction methods? (2) Will graph reduction methods make GNN more vulnerable to adversarial manipulation? (3) Can existing defense methods maintain their effectiveness when combined with graph reduction techniques? 
To answer these questions, we perform a comprehensive evaluation across four graph sparsification and six coarsening methods, diverse datasets, GNN architectures, and reduction ratios, comparing clean, poisoned, and reduced models. Our analysis provides a detailed and novel understanding of the interplay between graph reduction strategies and GNN robustness.
In summary, the contributions of this paper are as follows: 
\begin{itemize}
    \item \textbf{Graph Sparsification Reduces Adversarial Impact:} Graph sparsification effectively removes added poisoned edges during training, significantly mitigating the impact of poisoning attacks like Mettack \cite{mettack}. However, it is less effective against evasion attacks, such as PGD \cite{xu2019pgd}, as these occur during inference when sparsification no longer applies. 
    \item \textbf{Graph Coarsening Exacerbates Vulnerabilities:}
    While prior work~\cite{zhu2024robustness} demonstrates that graph coarsening is effective against backdoor attacks, oppositely, our findings reveal that graph coarsening methods amplify the impact of general poisoning attacks. Coarsening merges dissimilar nodes connected by poisoned edges into supernodes, creating noisy representations with high feature variance and incorrect labels. Additionally, unmerged poisoned edges persist, further degrading model performance.
    \item \textbf{Performance of GNN Defense under Graph Reduction:} When combined with graph sparsification, defensive GNNs retain or even improve their defense capabilities, providing strong protection against poisoning attacks. In contrast, coarsening disrupts these defenses by transferring edge perturbations into supernode structures, rendering robust GNN models less effective.
\end{itemize}
Our code is available at \url{https://github.com/RPI-DSPlab/Gnn_Reduction_Poisoning_Benchmark}.
\section{Related Work}

\subsection{Adversarial ML in Graph}
Various adversarial attacks against GNNs have been developed, with the aim of introducing subtle, often imperceptible, perturbations to the data and making the model to misclassify the inputs. A variety of taxonomies exist to classify these attacks, based on different criteria \cite{li2020deeprobust,jin2021categorization,zheng2021grb}: 

\noindent\textbf{Attack Phase.}~
Attacks can occur during either the training phase (i.e., poisoning attacks) or the inference phase (i.e., evasion attacks). Poison attacks \cite{mettack,xu2019pgd,bojchevski2019nea,geisler2021prbcd,dai2023unnoticeable,liu2023grad,li2022strg,zugner2020nettack,chen2018fga}  allow adversaries to modify the training graph in advance to mislead the learning process and produce a compromised model. Evasion attacks \cite{ma2022infmax,ju2023gia,chen2022agia,wen2024nodevoting}, on the other hand, perturb only the test data at inference time without affecting the model parameters. In this work, we focus on poisoning attacks, as they interact directly with graph reduction methods during training and often result in more severe model degradation. Nonetheless, we also include evasion attacks in our evaluation to provide a comprehensive understanding of how graph reduction influences adversarial robustness.

\noindent\textbf{Attack Target.}~
Adversarial goals can be either targeted—misclassifying a specific node or a small subset of nodes \cite{chen2018fga,zugner2020nettack}—or global, where the attack aims to degrade the model’s overall performance \cite{mettack,bojchevski2019nea,geisler2021prbcd,liu2023grad,li2022strg,xu2019pgd}. In this paper, we focus on global attacks to evaluate adversarial impacts in a graph-accelerated system, considering that training one global poisoned model is computationally more efficient than training multiple models targeting different individual nodes.

\noindent\textbf{Attack Strategy.}~
The most prevalent poisoning attacks are Graph Modification Attacks (GMA) \cite{mettack,bojchevski2019nea,geisler2021prbcd,liu2023grad,li2022strg,xu2019pgd,chen2018fga,zugner2020nettack}, where the attacker perturbs the graph topology by adding or removing edges. For example, in a social network, a user may connect or disconnect with others using existing accounts. Notably, GMAs can be used for both poisoning and evasion.
A special category of poisoning attacks is Graph Backdoor Attacks \cite{zhang2021backdoor,dai2023unnoticeable}, which implant a hidden trigger into the model by inserting specific patterns into the training data. The study by \cite{zhu2024robustness} investigates how graph reduction techniques affect the success of such backdoor attacks.
In the evasion setting, Graph Injection Attacks (GIA) \cite{chen2022agia,ju2023gia} introduce new malicious nodes with carefully crafted features that propagate misleading information to influence predictions during inference. Meanwhile, feature-based attacks \cite{ma2022infmax,wen2024nodevoting} directly manipulate node attributes to induce misclassification.


\subsection{Defense Methods}

A variety of defense strategies have been proposed for GNN robustness, which can broadly be categorized into preprocess-based \cite{entezari2020gcnsvd,wu2019jaccard} and model-based \cite{zhu2019robustgcn,zhang2020gnnguard,chen2021median} methods.
The preprocess-based methods aim to sanitize the graph before training by removing suspicious edges, whereas the model-based methods approaches focus on designing robust GNN architectures that penalize adversarial edges or nodes. In this study, we focus on model-based methods as they can be seamlessly integrated with accelerated systems. Preprocess-based methods often remove a significant number of edges, which may compromise the structural integrity and degrading performance—especially when applied with graph reduction.
\section{Preliminaries}

\subsection{Problem Definition}
An attributed graph $G=(V, E, X)$ is defined by (1) a set of $N$ nodes $V=\{ v_1, \ldots, v_N \}$; (2) a set of edges representing every pair of connections between nodes, where an edge between $v_i$ and $v_j$ can be defined as $(v_i, v_j, w)$ where $w$ represents the connection's weight; (3) a set of features $X \in \mathbb{R}^{N \times D}$, where D represents the feature dimension of each node. The node and edge sets can be written as a weighted adjacency matrix of $N$ nodes $A \in \mathbb{R}^{N \times N}$, where $A_{ij}=w$ if node $v_i$ and $v_j$ are connected with connection weight $w$. In a semi-supervised node-level classification task within an inductive setting, a small set of nodes $V_L \subseteq V$ in the graph are provided with labels $Y_L \in [0, 1]^{N_L\times C}$ where $C$ represents the number of classes. By feeding the subgraph $G_t = (V_t, E_t, X_t)$ that is composed of the nodes with known labels $Y_t$, a GNN model is trained to predict the class of unlabeled test nodes based on the cross-entropy loss function in Equation \ref{CrossEntropy}, where $f_\theta(\cdot)$ is the predict function with the use of GNN and $Y_c$ represents the ground truth labels' probability distribution of class $c$. 
\begin{equation}
\mathcal{L}(f_\theta(G_t)) = \sum_{c=1}^{C}Y_c \,\log f_\theta(G_t)_c\label{CrossEntropy}
\end{equation}

\noindent\textbf{Attack Models.}~
Adversarial attacks in GNNs are implemented by making a small number of perturbations $n_p$ to the edges and features, such as adding and removing an edge to mislead the model to classify a certain part of test nodes $V_p \subseteq V$ incorrectly. The poisoned graph is denoted as $G_p = (V, E_p, X_p)$, which is generated by minimizing the attack objective shown in Equation \ref{AttackLoss} where $l_{atk}$ is the loss function for the attack and $y_u$ is the label of node $u$. 
\begin{equation}
min \, \mathcal{L}_{atk}(f_\theta(G_p)) = \sum_{u \in V_p} l_{atk}(f_{\theta}(G_p)_u, y_u) \label{AttackLoss}
\end{equation}
We can further split the edges in the poisoned graph into three parts: original edges $E$, newly added edges $E_{add}$, and removed edges $E_{remove}$ followed by the equation \ref{Edges}. With original edges as the foundation, a poisoned graph can be extracted into two partially poisoned graphs with adding-only and remove-only strategies, denoted as $G_{p-add}$ and $G_{p-remove}$.
\begin{equation}
E_p = E + E_{add} - E_{remove} \label{Edges}
\end{equation}


For GIAs, the $G_p$ is generated by injecting a set of malicious nodes $V_{p}$ as 
$$
X_p = 
\begin{bmatrix}
X \\ 
X_{\text{atk}}
\end{bmatrix},
\;
A_p = 
\begin{bmatrix}
A & A_{\text{atk}} \\[6pt]
A_{\text{atk}}^T & O_{\text{atk}}
\end{bmatrix},
$$
where $X_{atk}$ is the features of the injected nodes, $O_{atk}$ is the adjacency matrix among injected nodes, and $A_{atk}$ is the adjacency matrix between the injected nodes and the original nodes.

\noindent\textbf{Attacker's Capability.}~ To match the experiment setting with real-world circumstances, we make the black-box assumption for global attacks. Specifically, attackers can view and edit the graph-structure data that they want to attack, including all nodes, features, edges, and labels. However, they do not have knowledge about the methods that defenders use, such as the GNN architecture. Following this assumption, to attack a GNN model and(or) test the attack's performance, attackers must train a surrogate model under a certain GNN architecture they presume to implement attacks. More than that, attackers should implement as little perturbation as possible to make the attack unnoticeable. 

\subsection{Graph Reduction}
The most time-expensive part of the GNN training algorithms comes from aggregation messages from neighboring nodes, which leads to a tremendous computation graph\cite{zhang2023survey}. Therefore, the primary goal of graph reduction methods is to modify the graph to be a smaller graph $G_r = (V_r, E_r, X_r)$, where $|V_r| < |V|$ and(or) $|E_r| < |E|$, with matched label set $Y_r$. The graph reduction function is $f_{reduced}(G, Y) = G_r, Y_r$. 

The graph coarsening method, followed by the framework in \cite{loukas2019graph}, partitioned a graph $G$ into $K$ clusters first, then a matrix $C \in \mathbf{R}^{N \times K}$ is used to represent the partition, which will be used to construct the super-nodes and super-edges to form a coarsen graph $G_r$. Features of super-nodes $X_r$ are set to be the weighted average of node features within each cluster. Similarly, the label set $Y_r$ is generated by the dominant label of nodes in each super-node, i.e., $Y_r = arg\,max(C^TY)$. 

Graph sparsification, on the other hand, aims to accelerate GNN training by reducing the graph size while preserving its structural and predictive properties. This is achieved by removing redundant edges while keeping the node features and labels unchanged. Various edge selection strategies have been explored to maintain similar classification performance, including random selection, degree-based selection \cite{hamann2016structure}, node-similarity-based selection \cite{satuluri2011local}, and edge-similarity-based selection \cite{xu2007scan}.
\section{Experiment}
\subsection{Experimental Settings}
\noindent\textbf{Dataset.}~
We use three datasets of different scales for GNN node classification that are commonly used to check adversarial attack performance: Cora, Pubmed, and CS, where Cora is a small graph with about 1000 nodes, Pubmed and CS are large graphs with about 10000 nodes. 
Detailed statistics of these datasets are provided in the supplementary Table \ref{tab:datasets}. 


\noindent\textbf{GNN Reduction Methods.}~
For the graph coarsening, we implement six methods that include three methods, Variation neighborhoods (VN), Variation edges (VE), and Variation climes (VC) from \cite{loukas2019graph}, Heavy Edge Matching (HE) \cite{loukas2018spectrally}, Algebraic JC (JC) \cite{ron2011relaxation}, and Kron (KRON) \cite{dorfler2012kron}. We define the reduction ratio of graph coarsening as the ratio of the number of nodes in the reduced graph $G_r$ to the original node number i.e., $r=\frac{|V_r|}{|V|}$, to indicate the portion of nodes that are merged into super-nodes during the coarsening. 

For graph sparsification, we implement four algorithms based on Networkit \cite{staudt2016networkit}: Random Node Edge (RNE), Local Degree (LD)\cite{hamann2016structure}, Local Similarity (LS)\cite{satuluri2011local}, and Scan (SCAN)\cite{xu2007scan}. Similar to the coarsening ratio, we define the reduction ratio of graph sparsification as the ratio of the number of edges in the sparsified graph $G_r$ to the number of edges in G to quantify how many edges are removed in this phase, i.e., $r=\frac{|E_r|}{|E|}$.

The detailed description of each coarsening and sparsification algorithm can be found in the supplementary Section \ref{sec:Appendix_A.1}.

\noindent\textbf{Attack models.}~
We implement seven most-recent global poisoning attacks, namely, DICE\cite{mettack}, NEA\cite{bojchevski2019nea}, PGD\cite{xu2019pgd}, Mettack\cite{mettack},  PRBCD\cite{geisler2021prbcd}, GraD\cite{liu2023grad}, and STRG-heuristic\cite{li2022strg} with the use of DeepRobust \cite{li2020deeprobust}, an open source benchmark package for adversarial attacks. We implement these attacks in a black-box setting, in which attackers do not have preliminary knowledge of which GNN model will be used for the training. We further constraint the total number of edges it can modify with a perturbation ratio $p=\frac{|E_{add}|+|E_{remove}|}{E}$.

\noindent\textbf{GNN Models.}~
We consider three common GNN architectures, namely, GCN\cite{kipf2016semi}, GraphSAGE \cite{hamilton2017inductive}, GAT \cite{velivckovic2017graph}, and three powerful defensive GNN models, Robust GCN(RGCN)\cite{zhu2019robustgcn}, GNNGuard\cite{zhang2020gnnguard}, and MedianGCN\cite{chen2021median}. 

\noindent\textbf{Evaluation Metrics.}~
We evaluate the classification performance on the test set sample from the global graph by calculating the clean accuracy($ACC_c$) using the clean adjacency matrix to train, poisoned accuracy($ACC_p$) using the poisoned adjacency matrix, and post-reduction accuracy($ACC_r$) that uses the reduced poisoned graph. 

\subsection{Baseline Effectiveness Analysis}
Before testing the effect of poisoning attacks under graph reductions, we systematically evaluate each attack's performance under different GNN architectures. We first use a clean adjacency matrix and clean feature to train a surrogate GCN model. This surrogate model will be used by attackers to implement and test their perturbations. Each experiment was repeated five times under different random seeds and then averaged the clean accuracy and poisoned accuracy. 

Table \ref{tab:PoisonAttacks} shows the clean accuracy $ACC_c$ and poisoned accuracy $ACC_p$ of GCN trained on three datasets under different perturbation ratios (2\%, 5\%, 10\%). The results indicate that all attacks reduce model accuracy, with Mettack and GraD leading in degrading performance under a limited perturbation budget. To align with the attacks' goal to ensure attack impact with minimal perturbation, we use a 5\% perturbation ratio as our hyperparameter in the experiments below as this level consistently produces a notable drop in prediction accuracy across multiple attacks.

\begin{table}
\centering
\caption{GCN clean accuracy and poisoned accuracy under five poisoning attacks with various perturbation ratios. The lowest poisoned accuracy is \textbf{highlighted} in each dataset. N/A indicates the experiment is out of memory.}
\begin{tabular}{cr|rrr} 
\toprule
\multicolumn{1}{l}{}  & \multicolumn{1}{l|}{} & \multicolumn{3}{c}{$ACC_p$}\\
\multicolumn{1}{l}{Dataset ($ACC_c$)} & \multicolumn{1}{l|}{Attack} & $p$=2\%  & 5\% & 10\%\\
\hline
\multirow{7}{*}{Cora (84.06\%)}    
    & DICE         & 83.40\% & 81.98\% & 80.62\% \\
    & NEA          & 83.20\% & 82.46\% & 81.25\% \\
    & PGD          & 81.71\% & 78.20\% & 74.77\% \\
    & Mettack      & \textbf{81.26\%} & \textbf{74.46\%} & \textbf{71.22\%} \\
    & PRBCD        & 83.23\% & 81.11\% & 79.68\% \\
    & STRG-Heuristic & 81.94\% & 79.07\% & 77.19\%\\
    & GraD         & 80.56\% & 76.26\% & 72.52\%\\
\hline
\multirow{7}{*}{Pubmed (86.15\%)}    
    & DICE          & 85.46\% & 85.21\% & 83.40\% \\
    & NEA      & 85.63\% & 85.49\% & 84.30\% \\
    & PGD          & \textbf{84.12\%} & 81.33\% & 78.08\% \\
    & Mettack      & 84.95\% & \textbf{78.07\%} & \textbf{62.78\%} \\
    & PRBCD        & 84.77\% & 83.39\% & 81.53\% \\
    & STRG-Heuristic & 85.33\% & 84.73\% & 82.94\%\\
    & GraD         & 82.17\% & 77.72\% & 65.93\%\\
\hline
\multirow{7}{*}{CS (92.43\%)}    
    & DICE          & 92.20\% & 91.81\% & 91.15\% \\
    & NEA      & 92.20\% & 91.85\% & 91.35\% \\
    & PGD          & 90.76\% & 88.71\% & 86.37\% \\
    & Mettack      & \textbf{88.49\%} & \textbf{82.25\%} & \textbf{76.95\%} \\
    & PRBCD        & 90.72\% & 88.95\% & 86.89\% \\
    & STRG-Heuristic & 92.14\% & 91.78\% & 91.06\%\\
    & GraD         & N/A & N/A & N/A\\
\hline
\end{tabular}
\label{tab:PoisonAttacks}
\end{table}

We further evaluate the robustness of different GNN models and defense methods against poisoning attacks with a perturbation ratio of 5\% in Table \ref{tab:PoisonAttackGNNs}, where GNNs trained with poisoned graph data that have accuracy close to clean models within the 1\% range are highlighted. As shown in the Table, every attack effectively decreased the classification accuracy in small datasets. However, in large datasets, DICE, NEA, and STRG-Heuristic reflect marginal attacking performance, while other attacks remain powerful. Among these attacks, Mettack and GraD show the strongest attacking performance with the lowest poisoned accuracy. From the defender's perspective, GNNGuard consistently mitigates most attacks on large datasets with minimal accuracy loss.

\begin{table}
\centering
\scriptsize
\caption{Poisoned accuracy under five poisoning attacks with 5\% perturbation ratios trained by GNNs. The highest accuracy in each attack is \textbf{bolded}.}
\label{tab:PoisonAttackGNNs}
\setlength{\tabcolsep}{2pt} 
\begin{tabular}{cr|rrrrrr} 
\toprule
\multicolumn{1}{l}{}  & \multicolumn{1}{l|}{} & \multicolumn{6}{c}{$ACC_p$}\\
\multicolumn{1}{l}{Dataset}  & \multicolumn{1}{l|}{Attack} & GCN & GAT & SAGE & GNNGuard & RGCN & Median\\
\hline
\multirow{8}{*}{Cora}
    & Clean        & 84.06\% & \textbf{84.55\%} & 83.63\% & 80.59\% & 83.82\% & 84.52\%\\
    & DICE         & 81.98\% & 81.41\% & 82.01\% & 77.59\%& 81.73\% &               \textbf{83.15}\%\\
    & NEA          & 82.46\% & 82.70\% & 81.79\% & 79.06\% & 82.34\% &     \textbf{83.88}\%\\
    & PGD          & 78.20\% & 78.25\% & 78.52\% & 77.04\% & 78.23\% & \textbf{79.14\%}\\
    & Mettack      & 74.46\% & \textbf{77.83\%} & 76.68\% & 76.30\% & 73.75\% & 77.64\%\\
    & PRBCD        & 81.11\% & \textbf{81.84\%} & 80.78\% & 77.11\% & 80.46\% & 81.57\%\\
    & GraD         & 76.26\% & 80.86\% & 79.53\% & 78.40\% & 74.30\% & \textbf{81.30\%}\\
    & STRG-Heuristic& 79.07\% & 80.84\% & 79.32\% & 76.71\% & 77.89\% & \textbf{80.89\%}\\
\hline
\multirow{8}{*}{Pubmed}
    & Clean & \textbf{86.15\%} & 85.13\% & 85.59\% & 84.67\% & 85.49\% & 84.43\%\\
    & DICE  & 85.21\% & 83.83\% & \textbf{85.10\%} & 84.52\% & 84.63\% & 83.56\%\\
    & NEA   & 85.49\% & 84.50\% & \textbf{85.69\%} & 84.70\% & 85.17\% & 84.19\%\\
    & PGD   & 81.33\% & 80.67\% & 83.35\% & \textbf{83.53\%} & 80.70\% & 80.82\%\\
    & Mettack & 78.07\% & 82.95\% & 81.40\% & \textbf{84.67\%} & 81.80\% &     79.39\%\\
    & PRBCD & 83.39\% & 82.56\% & \textbf{85.08\%} & 84.20\% & 83.18\% & 83.33\%\\
    & GraD  & 77.72\% & 81.16\% & 80.23\% & \textbf{84.64\%} & 79.51\% & 82.52\%\\
    & STRG-Heuristic& 84.73\% & 83.81\% & \textbf{85.67\%} & 84.30\% & 84.46\% & 82.60\%\\
\hline
\multirow{8}{*}{CS}
    & Clean & 92.43\% & 92.10\% & 92.13\% & \textbf{92.60\%} & 92.08\% & 91.91\%\\
    & DICE  & 91.81\% & 91.56\% & 91.72\% & \textbf{92.36\%} & 91.28\% & 91.28\%\\
    & NEA   & 91.85\% & 90.47\% & 91.60\% & \textbf{92.53\%} & 91.02\% & 91.19\%\\
    & PGD   & 88.71\% & 88.78\% & 89.39\% & \textbf{91.51\%} & 88.33\% & 88.05\%\\
    & Mettack & 82.25\% & 89.26\% & 88.10\% & \textbf{92.25\%} & 89.02\% & 90.17\%\\
    & PRBCD & 88.95\% & 88.90\% & 90.97\% & \textbf{92.31\%} & 88.68\% & 89.58\%\\
    & GraD  & N/A & N/A & N/A & N/A & N/A & N/A\\
    & STRG-Heuristic& 91.78\% & 91.66\% & 92.21\% & \textbf{92.49\%} & 89.44\% & 90.83\%\\
\hline
\end{tabular}
\end{table}
\subsection{Robustness Under Graph Reduction}
In this section, we empirically evaluate the effect of graph coarsening and sparsification on poisoning attacks. After the poisoning pipeline, the perturbed graph is passed to the graph coarsening/sparsification algorithm to reduce its size with various reduction ratios between 0 and 1. The reduced poisoned graph is then fed into the GNN model to train and test its accuracy and robustness.

Table \ref{tab:reduced_acc} shows the GCN's poisoned accuracy $ACC_p$ for each attack and its corresponding post-reduction accuracy $ACC_r$.
After graph sparsification, some attacks, e.g., PGD and PRBCD, maintain similar attack performance, while other attacks, e.g., Mettack and GraD, are largely mitigated by graph sparsification, leading to an increased $ACC_r$. 

In contrast, graph coarsening amplifies adversarial influence by further reducing accuracy compared with graph sparsification. The worst case is Mettack, where the average $ACC_r$ is $61.65\%$, which is $16.41\%$ lower than the Mettack attack's accuracy under no graph reduction. 
\begin{table*}[t]
\footnotesize
\setlength{\tabcolsep}{2pt} 
\centering
\caption{GCN's clean accuracy, poisoned accuracy under $p=0.05\%$, and post-reduction accuracy after graph coarsening and sparsification under $r=0.325$. The $ACC_r$ with lower than 1\% drop compared to $ACC_c$ is bolded. }
\begin{tabular}{cr|r|rrrr|rrrrrr}
\toprule
\multicolumn{1}{l}{}  & \multicolumn{1}{l|}{} & \multicolumn{1}{l|}{} & \multicolumn{10}{c}{$ACC_r$}\\
Dataset ($ACC_c$) & Attack & $ACC_p$ & RNE & LD & LS & SCAN & VN & VE & VC & HE & JC & KRON\\
\hline
\multirow{7}{*}{Cora (84.06\%)}
& DICE    & 81.98\%
    & 80.63\% & 81.22\% & 80.51\% & 81.06\%
    & 80.42\% & 79.44\% & 82.14\% & 78.66\% & 79.70\% & 79.98\%\\
& NEA     & 82.46\% 
    & 81.49\% & \textbf{83.50\%} & 81.67\% & 79.47\%
    & 81.55\% & 79.37\% & 80.12\% & 79.89\% & 79.30\% & 80.17\%\\
& PGD     & 78.20\% 
    & 77.88\% & 78.91\% & 78.94\% & 76.34\%
    & 76.41\% & 75.51\% & 79.62\% & 77.37\% & 78.40\% & 76.24\%\\
& Mettack & 74.46\% 
    & 79.22\% & 80.17\% & 76.50\% & 80.23\%
    & 78.34\% & 80.00\% & 77.92\% & 77.70\% & 75.17\% & 77.82\%\\
& PRBCD   & 81.11\% 
    & 80.30\% & 81.24\% & 81.16\% & 81.10\%
    & 81.09\% & 79.74\% & 79.69\% & 79.16\% & 79.76\% & 78.88\%\\
& GraD   & 76.26\% 
    & 80.22\% & 80.50\% & 78.90\% & 80.99\% 
    & 80.38\% & 79.28\% & 79.58\% & 79.15\% & 80.20\% & 76.42\%\\
& STRG-Heuristic   & 79.07\% 
    & 78.40\% & 80.55\% & 78.40\% & 79.44\% 
    & 78.51\% & 77.09\% & 75.04\% & 75.07\% & 78.17\% & 75.07\%\\
\hline
\multirow{7}{*}{Pubmed (86.15\%)}
& DICE    & 85.21\% 
    & 84.49\% & 84.94\% & 84.59\% & 84.91\%
    & 83.34\% & 82.72\% & 82.62\% & 82.44\% & 82.67\% & 82.87\%\\
& NEA     & 85.49\% 
    & 84.78\% & 85.24\% & 85.40\% & 85.28\%
    & 83.47\% & 82.38\% & 83.26\% & 83.01\% & 83.28\% & 84.28\%\\
& PGD     & 81.33\% 
    & 81.23\% & 81.24\% & 81.46\% & 81.34\%
    & 79.91\% & 79.46\% & 78.78\% & 79.19\% & 79.70\% & 79.45\%\\
& Mettack & 78.07\% 
    & \textbf{85.80\%} & \textbf{86.27\%} & 79.78\% & 64.19\%
    & 72.54\% & 61.76\% & 64.18\% & 45.27\% & 61.97\% & 64.21\%\\
& PRBCD   & 83.39\% 
    & 82.97\% & 83.95\% & 83.45\% & 83.05\%
    & 82.01\% & 81.00\% & 81.68\% & 81.87\% & 78.88\% & 82.05\%\\
& GraD   & 77.72\% 
    & \textbf{85.34\%} & \textbf{85.28\%} & 82.32\% & 76.93\% 
    & 80.95\% & 75.29\% & 74.07\% & 68.78\% & 75.41\% & 73.52\%\\
& STRG-Heuristic  & 84.73\%
    & 84.98\% & \textbf{85.15\%} & 83.56\% & 82.13\% 
    & 83.00\% & 78.68\% & 77.27\% & 77.77\% & 76.54\% & 79.64\%\\
\hline
\multirow{7}{*}{CS (92.43\%)}
& DICE    & 91.81\% 
    & 90.91\% & 90.16\% & 90.79\% & 90.68\%
    & 90.12\% & 89.82\% & 90.01\% & 89.89\% & 89.97\% & 89.91\%\\
& NEA     & 91.85\%
    & 90.69\% & 91.27\% & 90.80\% & 90.36\%
    & 90.48\% & 90.18\% & 89.79\% & 90.25\% & 89.81\% & 89.34\%\\
& PGD     & 88.71\%
    & 88.54\% & 88.60\% & 88.37\% & 88.15\%
    & 87.95\% & 87.29\% & 87.53\% & 87.40\% & 87.11\% & 86.68\%\\
& Mettack & 82.25\%
    & 91.32\% & 90.53\% & 90.49\% & 91.09\%
    & 84.80\% & 85.25\% & 83.57\% & 84.54\% & 83.94\% & 84.69\%\\
& PRBCD   & 88.95\% 
    & 88.16\% & 88.51\% & 87.77\% & 87.45\%
    & 87.36\% &87.50\% & 87.03\% & 87.46\% & 87.39\% & 87.06\%\\
& GraD   & N/A 
    & N/A & N/A & N/A & N/A 
    & N/A & N/A & N/A & N/A & N/A & N/A\\
& STRG-Heuristic   & 91.78\% 
    & 91.32\% & 91.37\% & 90.89\% & 90.46\% 
    & 88.73\% & 88.73\% & 88.13\% & 87.57\% & 86.63\% & 84.99\%\\
\hline
\end{tabular}
\label{tab:reduced_acc}
\end{table*}

\begin{figure}
    \centering
    \includegraphics[width=1\linewidth]{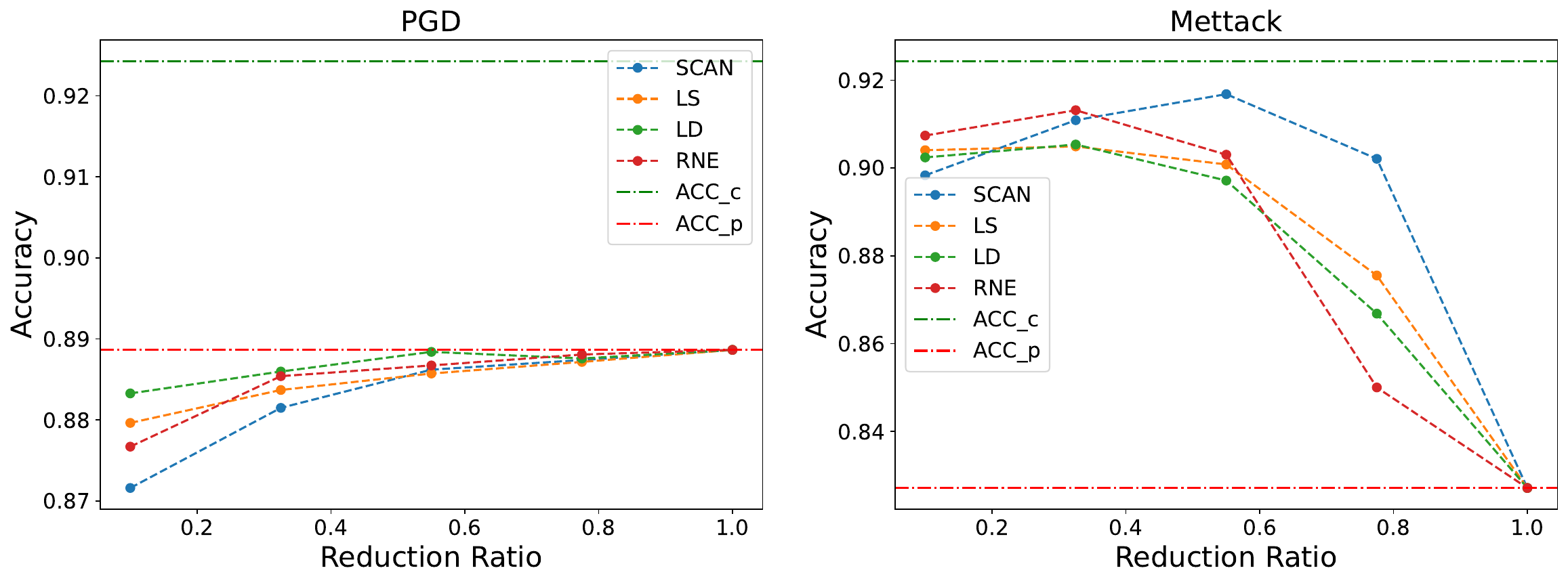}
    \caption{$ACC_r$ using four sparsification algorithms against PGD (left) and Mettack (right) with reduction ratios $r$ in CS dataset.}
    \label{fig:sparsification_compare}
\end{figure}

\begin{figure}
    \centering
    \includegraphics[width=1\linewidth]{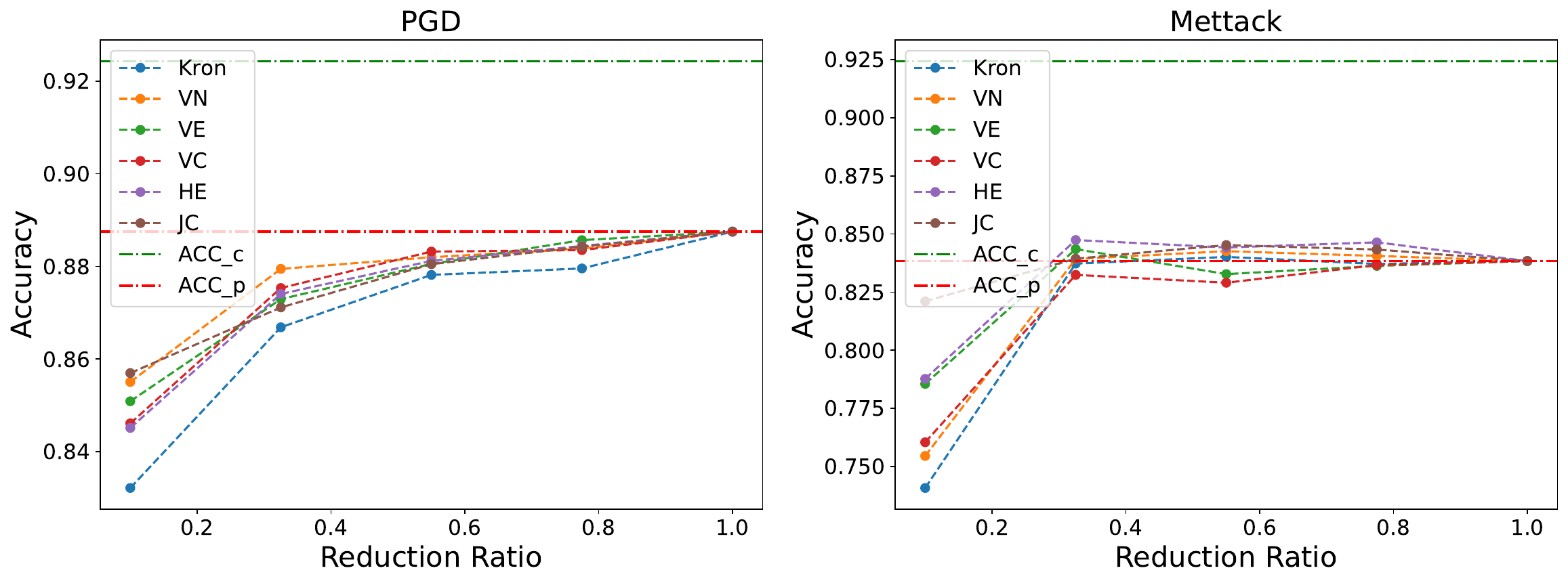}
    \caption{$ACC_r$ using six coarsening algorithms against PGD (left) and Mettack (right) with various $r$ in CS dataset.}
    \label{fig:coarsening_compare}
\end{figure}

We further investigate the impact of graph reduction algorithms on classification accuracy across varying reduction ratios. Using PGD and Mettack, we illustrate the effect of sparsification and coarsening under different reduction ratios in Figure \ref{fig:sparsification_compare} and Figure \ref{fig:coarsening_compare}, respectively. Complete figures for accuracy performance under graph reduction for every poisoning attack are provided in Appendix Figure \ref{fig:complete_coarsening_gcn} and Figure \ref{fig:complete_sparsification_gcn}. For graph sparsification, the reduction ratio has a minimal impact on attack performance under PGD attack, with only a slight decrease in accuracy due to the reduced training graph size. In contrast, under Mettack, for all sparsification methods, the accuracy dramatically increased as the reduction ratio decreased, indicating the mitigation effect of graph sparsification against Mettack. On the other hand, as shown in Figure \ref{fig:coarsening_compare}, graph coarsening after both attacks results in a sharp decline in accuracy as the reduction ratio decreases from 0.325 to 0.1, suggesting that graph coarsening amplifies the effectiveness of adversarial attacks under small reduction ratios.





\subsection{Perturbation Edge Analysis}
Given the observations in the previous section, we conduct a thorough edge analysis to understand why graph sparsification can effectively defend against some attacks while failing in others. Similarly, we want to explore why the performance of adversarial attacks can be amplified by graph coarsening.
\subsubsection{Graph Sparsification}
\label{sssec:sparsification}
\begin{figure}
    \centering
    \includegraphics[width=1\linewidth]{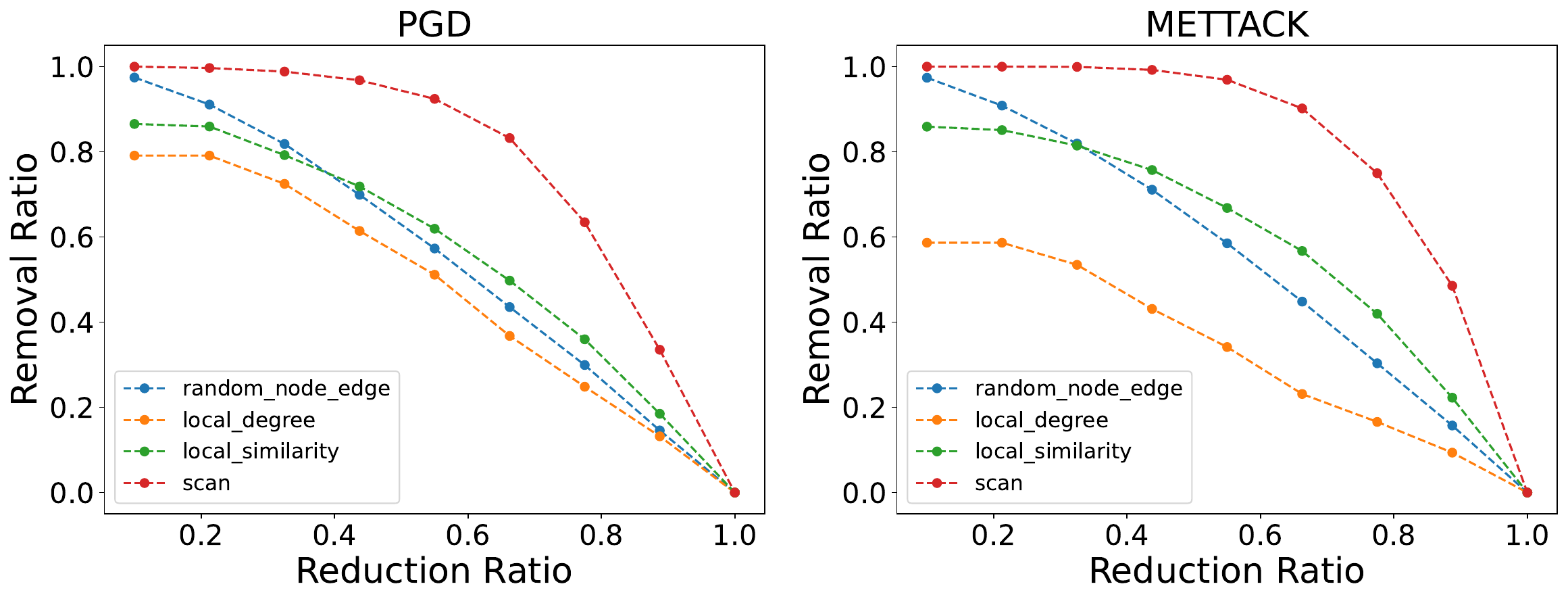}
    \caption{The ratio of removed newly-added perturbation edges by PGD(left) and Mettack(right) after four sparsification methods with various $r$ in the CS dataset.}
    \label{fig:merge_compare}
\end{figure}

For graph sparsification, we evaluate the proportion of poisoned edges removed during the reduction process. Figure \ref{fig:merge_compare} illustrates the removal ratios under varying sparsification levels ($r_s$) for both the PGD, which retains the attacking performance after sparsification, and Mettack, which is largely mitigated by sparsification.
A complete edge analysis result for every attack in every dataset can be found in Appendix Figure \ref{fig:complete_merge_compare}.
Surprisingly, for both attacks, sparsification methods' removal ratio of poisoned edges is about or above $60\%$, indicating most of the poisoned edges are removed in the training phase after graph sparsification. Given such observation, we propose two hypotheses to explain why certain poisoning attacks retain their adversarial influence despite sparsification: (1) These attacks primarily remove existing edges rather than add new ones, reducing the impact of sparsification on their effectiveness; (2) Although graph sparsification removes a significant portion of poisoned edges during training, resulting in a cleaner model, adversarial perturbations can still take effect during the inference phase as evasion attacks.
To identify the underlying reasons, we analyze every component of the attacks and their respective impact on the classification accuracy in Table \ref{tab:EdgeRatio}. On the first half of the table, we observe that all optimization-based attacks predominantly focus on adding edges rather than removing them. The exception is the rule-based attack DICE, which explicitly follows a predefined rule to externally connect edges, resulting in a balance between added and removed edges. Such observation excludes hypothesis one. 

\begin{table}
\centering
\caption{Each attack's added edge ratio and removal edge Ratio, as well as its attacking performance under both poison and evasion setting.}
\begin{tabular}{c|cc|cc}
\multicolumn{1}{c|}{} & \multicolumn{2}{c|}{} & \multicolumn{2}{c}{\centering$ACC_p$}\\
Attack & $E_{add} / E_p$ & $E_{remove} / E_p$ & $G_p$ & $G$(evasion)\\
\hline
DICE & 50.20\% & 49.80\% & 84.89\% & 85.03\%\\
NEA & 100\% & 0\% & 85.42\% & 85.50\%\\
PGD & 96.66\% & 3.34\% & 81.33\% & 81.40\%\\
Mettack & 100\% & 0\% & 78.07\% & 86.,67\%\\
PRBCD & 100\% & 0\% & 83.39\% & 82.74\%\\
GraD & 100\% & 0\% & 77.42\% & 85.94\%\\
STRG-Heuristic & 100\% & 0\% & 83.99\% & 84.89\%\\
\hline
\end{tabular}
\label{tab:EdgeRatio}
\end{table}


On the second half of the table, we analyze the effectiveness of each perturbation component by evaluating the GCN model's inference accuracy on the poisoned graph $G_p$ with models trained on $G_p$ and $G$. Note that training the model with a clean graph but inference using a poisoned graph is an evasion attack setting. We observe that while Mettack effectively poisons a GNN during the training phase, its adversarial impact diminishes in the evasion attack setting, where perturbations are introduced only during inference. In contrast, PGD and PRBCD retain their adversarial effectiveness in both poisoning and evasion attack scenarios, indicating that their perturbations remain impactful regardless of whether the model is trained on a poisoned graph or not. Therefore, although graph reduction methods eliminated most of the poisoned edges in the training phase, resulting in a clean GNN model, because of the nature of semi-supervised learning, the poisoned graph will still be used in the inference phase, which causes a degradation in classification accuracy.

\subsubsection{Graph Coarsening}
\label{sssec:coarsening}
\begin{figure}
    \centering
    \includegraphics[width=1\linewidth]{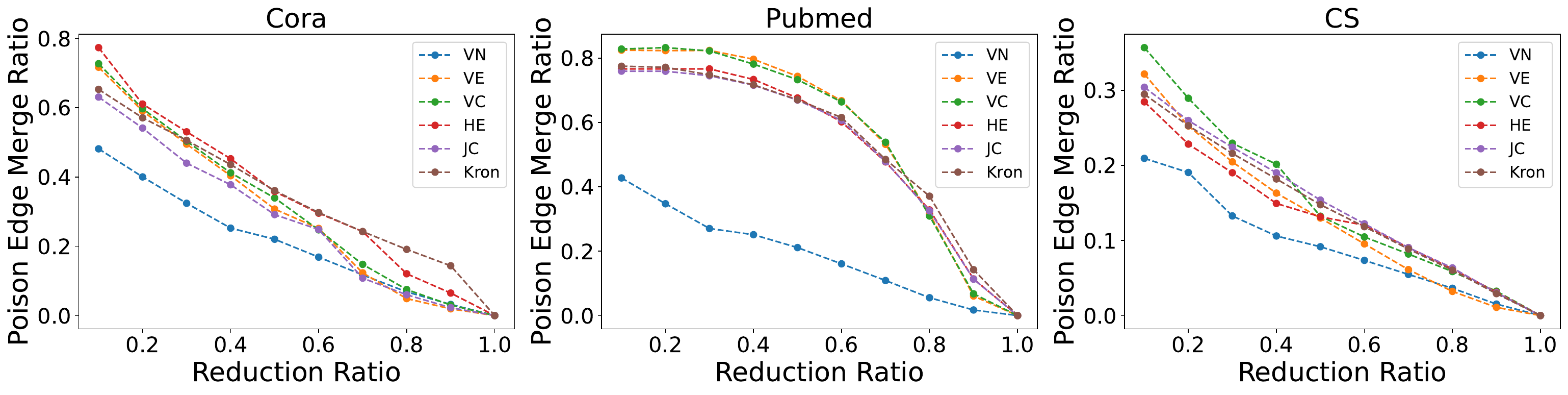}
    \caption{The merge ratio of newly-added perturbation edges generated by PGD after six coarsening methods with various reduction ratios $r$.}
    \label{fig:coarsening_merge_compare_pgd}
\end{figure}

\begin{figure}
    \centering
    \includegraphics[width=1\linewidth]{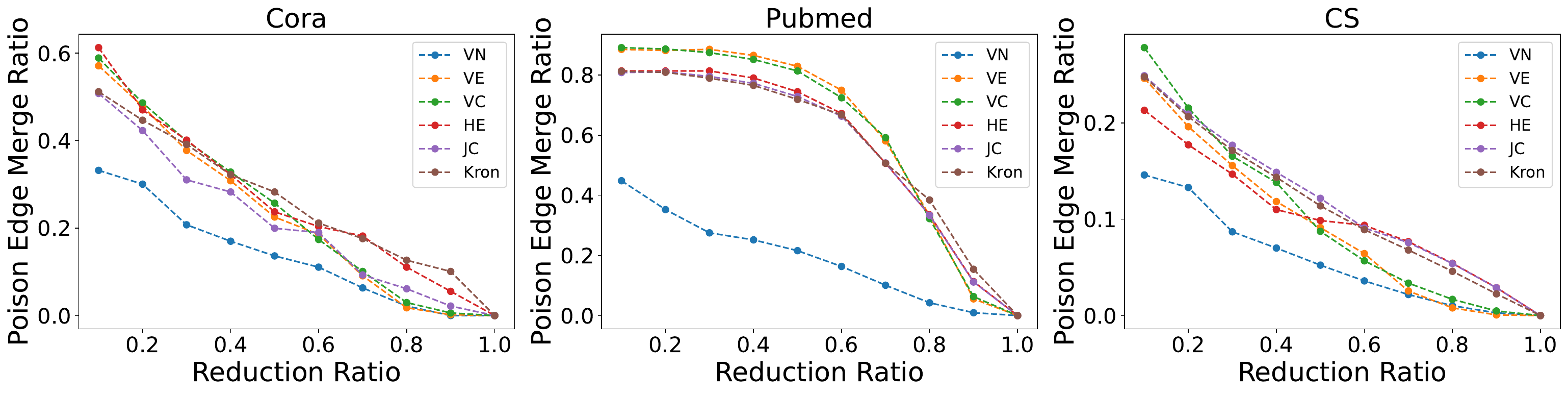}
    \caption{The merge ratio of newly-added perturbation edges generated by Mettack after six coarsening methods with various reduction ratios $r$.}
    \label{fig:coarsening_merge_compare_mettack}
\end{figure}

For graph coarsening, similarly, we evaluate the merge ratio, which represents the ratio of malicious edges added by a poisoning attack being removed and merged into super-nodes during coarsening. As shown in Figure \ref{fig:coarsening_merge_compare_pgd} and \ref{fig:coarsening_merge_compare_mettack}, unlike sparsification, which can effectively remove most of the added poisoning edges in every dataset, graph coarsening can only remove a small portion of such perturbations in the CS dataset(e.g., $35.69\%$ for PGD attack and $27.82\%$ for Mettack). In such cases, a vast number of poisoning edges will maintain the adversarial influence on the model during the training process.


\begin{figure}
    \centering
    \includegraphics[width=1\linewidth]{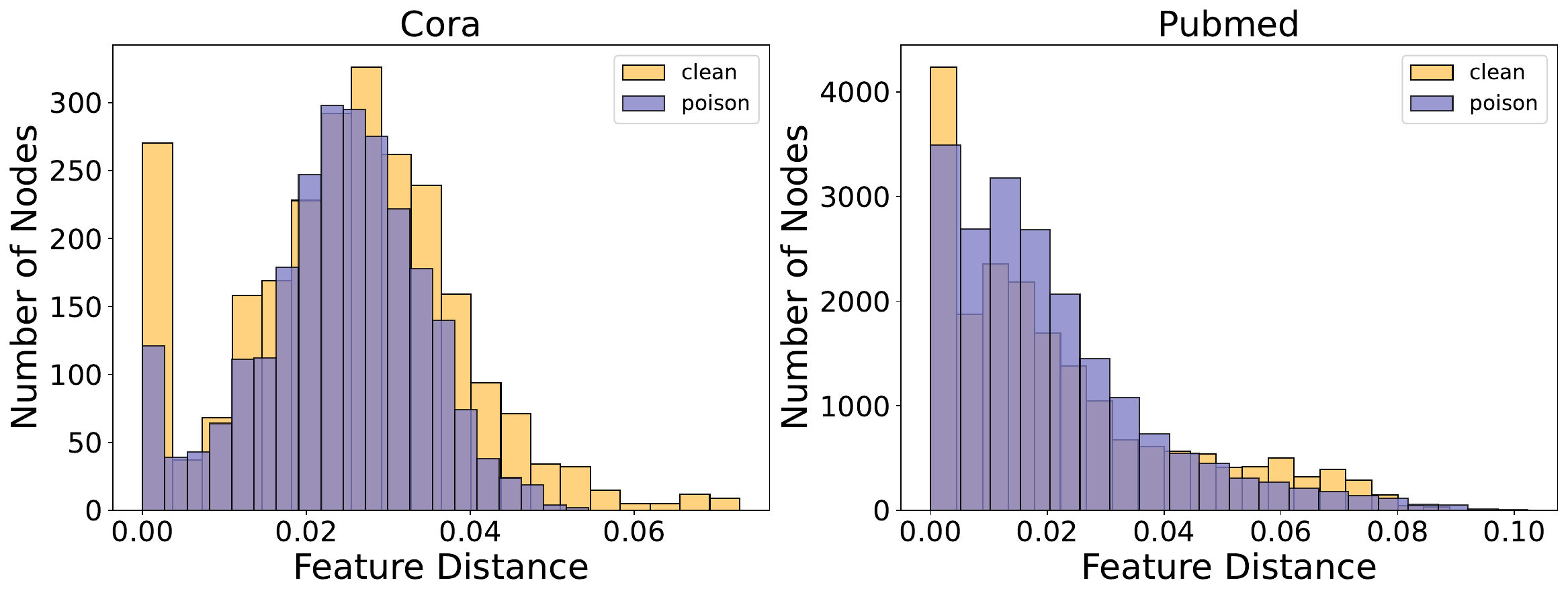}
    \caption{Comparison of feature distances' distribution on the Cora dataset (left) and the Pubmed dataset (right) between each node's feature and its corresponding super node's feature after graph coarsening on the clean graph (yellow) and the poisoned graph (blue).}
    \label{fig:feature_distance_compare}
\end{figure}

On the other hand, for the Cora and Pubmed datasets, although a large portion of poisoned edges are eliminated or merged into supernodes during the Mettack attack, it retains a strong adversarial impact, and graph coarsening does not improve classification accuracy by removing these perturbations. This suggests that supernodes inherit the adversarial effects of merged poisoned edges. To quantify such inherited perturbation, we evaluate (1) the distribution of Euclidean distance between each node's original feature and its corresponding super node's feature, and (2) the label difference ratio between the original graph and the coarsened graph, which measures the portion of nodes in the original graph whose labels are different from their corresponding super nodes' labels. As illustrated in Figure \ref{fig:feature_distance_compare}, the feature distance distribution for both clean and poisoned graphs (generated by Mettack) on the Cora and Pubmed datasets shows a marked decrease in the number of low-distance nodes. Our extended version \cite{wu2024understanding} provides full feature distance distributions for each dataset and attack. As illustrated in the Figure, the feature distance distribution for both clean and poisoned graphs (generated by Mettack) on the Cora and Pubmed datasets shows a marked decrease in the number of low-distance nodes. 

Table \ref{tab:label_diff} presents the label difference ratio results for the same datasets and poisoning attack, revealing an increased discrepancy between the labels of original nodes and their corresponding supernodes after the poisoning attack, compared to the clean coarsened graph. The complete results for all datasets and poisoning attacks are provided in supplementary Figure \ref{fig:complete_coarsening_gcn} and \ref{fig:complete_sparsification_gcn}. Together, these results suggest that applying graph coarsening after a poisoning attack leads to an unclean coarsened graph where original nodes are poorly matched with their supernodes—characterized by large feature distances—which in turn exacerbates the rate of misclassification.
\begin{table}
\centering
\caption{Label difference ratio between original node's label and its corresponding super node's label by VC coarsening with $r=0.3$ against PGD and Mettack}
\begin{tabular}{cr|r|r}
\toprule
\multicolumn{1}{l}{}  & \multicolumn{1}{l|}{} & \multicolumn{2}{l}{Label Diff Ratio}\\
Attack & Dataset & Clean & Poisoned\\

\hline
\multirow{3}{*}{PGD}
& Cora & 42.95\% & 44.83\%\\
& Pubmed & 42.75\% & 41.25\%\\
& CS & 45.44\% & 47.08\%\\
\hline
\multirow{3}{*}{Mettack}
& Cora & 40.72\% & 42.37\%\\
& Pubmed & 42.53\% & 53.02\%\\
& CS & 45.21\% & 49.47\%\\
\hline
\end{tabular}
\label{tab:label_diff}
\end{table}
\begin{table*}[ht!]
\footnotesize
\centering
\caption{Pubmed's clean accuracy, poisoned accuracy under $p=0.05\%$, and reduced accuracy after graph reduction under $r=0.325$ with various GNN architecture. N/A indicates that RGCN cannot be applied with graph coarsening as it requires the training graph to have the same number of nodes as the test graph.
}
\setlength{\tabcolsep}{2pt} 
\begin{tabular}{cr|rr|rrrr|rrrrrr}
\toprule
\multicolumn{1}{l}{}   & \multicolumn{1}{l|}{} & \multicolumn{1}{l}{} & \multicolumn{1}{l|}{} & \multicolumn{10}{c}{$ACC_r$}\\
\multicolumn{1}{l}{Attack}  & \multicolumn{1}{l|}{GNN} & $ACC_c$ & \multicolumn{1}{l|}{$ACC_p$} & RNE & LD & LS & SCAN & VN & VE & VC & HE & JC & KRON\\
\hline
\multirow{6}{*}{PGD}
& GCN & 86.15\% & 81.33\% 
    & 81.23\% & 81.17\% & 81.46\% & 81.34\% 
    & 79.91\% & 79.46\% & 78.78\% & 79.19\% & 79.70\% & 79.45\%\\
& GAT & 85.13\% & 80.67\%
    & 80.36\% & 80.78\% & 80.51\% & 80.11\%
    & 79.61\% & 73.47\% & 74.97\% & 78.51\% & 78.25\% & 78.84\%\\
& SAGE & 85.59\% & 83.35\%
    & 83.89\% & 82.86\% & 83.74\% & \textbf{84.67\%}
    & 79.48\% & 75.09\% & 75.32\% & 77.47\% & 77.82\% & 76.91\%\\
& GNNGuard & 84.67\% & \textbf{83.53\%}
    & 83.63\% & 83.53\% & \textbf{83.70\%} & \textbf{84.19\%}
    & 82.57\% & 81.32\% & 81.06\% & 81.17\% & 81.52\% & 81.80\%\\
& RGCN & 85.49\% & 80.70\%
    & 80.48\% & 80.77\% & 80.74\% & 80.46\%
    & N/A & N/A & N/A & N/A & N/A & N/A\\
& Median & 84.43\% & 80.82\%
    & 80.62\% & 81.12\% & 80.70\% & 80.94\%
    & 80.14\% & 78.22\% & 78.64\% & 78.45\% & 79.18\% & 79.70\%\\
\hline
\multirow{6}{*}{Mettack}
& GCN & 86.15\% & 78.07\%
    & \textbf{85.80\%} & \textbf{86.27\%} & 79.78\% & 64.18\%
    & 64.20\% & 59.41\% & 61.76\% & 45.27\% & 61.97\% & 64.21\%\\
& GAT & 85.13\% & 82.95\%
    & 83.93\% & \textbf{84.47\%} & 79.66\% & 64.84\%
    & 63.07\% & 58.94\% & 55.65\% & 50.72\% & 57.64\% & 63.07\%\\
& SAGE & 85.59\% & 81.40\%
    & \textbf{85.36\%} & \textbf{86.21\%} & 83.28\% & 74.56\%
    & 62.39\% & 54.26\% & 55.58\% & 45.41\% & 54.83\% & 62.39\%\\
& GNNGuard & 84.67\% & \textbf{84.67\%}
    & \textbf{84.54\%} & \textbf{84.80\%} & \textbf{84.60\%} & \textbf{84.65\%}
    & 63.64\% & 61.48\% & 60.69\% & 48.01\% & 62.26\% & 63.64\%\\
& RGCN & 85.49\% & 81.80\%
    & \textbf{85.19\%} & \textbf{85.17\%} & 81.28\% & 80.67\%
    & N/A & N/A & N/A & N/A & N/A & N/A\\
& Median & 84.43\% & 79.39\%
    & \textbf{83.97\%} & \textbf{84.12\%} & 78.09\% & 67.04\%
    & 62.01\% & 57.65\% & 61.41\% & 46.91\% & 60.34\% & 62.01\%\\
\hline
\label{tab:pubmed_gnns_reduced_acc}
\end{tabular}
\end{table*}
\subsection{Graph Reduction with Defensive GNNs}
In this section, we further look into the question of how the performance of robust GNN architectures is affected when they are deployed in a graph-accelerated system using graph reduction. For each poisoning attack, we compare the poisoned accuracy of defensive models across different graph reduction methods. Table \ref{tab:pubmed_gnns_reduced_acc} shows the accuracy of GNN models trained on Pubmed graphs that were first poisoned by poisoning attacks and then pre-processed by graph reduction. The full table for every dataset and every poisoning attack can be found in Appendix Table \ref{tab:complete_cora_gnns_reduced_acc}, \ref{tab:complete_pubmed_gnns_reduced_acc}, and \ref{tab:complete_cs_gnns_reduced_acc}.

As we can see, defensive models including GNNGuard, RGCN, and Median retain similar performance after graph sparsification with a slight fluctuation. For example, under PGD attack, the accuracy changes from $ACC_p = 83.53\%$ to $ACC_r = 84.19\%$ using GNNGuard with the SCAN algorithm and from $ACC_p = 80.82\%$ to $ACC_r = 80.94\%$ using MedianGCN with SCAN. In contrast, as we discussed in Section \ref{sssec:coarsening}, because graph coarsening inherits the perturbation edges into unclean super nodes that carry poisoned features, originally effective defensive GNNs like GNNGuard that improve robustness by eliminating suspicious edges, cannot retrain the same defense strength. For example, using GNNguard with HE coarsening, the accuracy changes from $ACC_p = 84.67\%$ to $ACC_r = 48.01\%$, decreasing by 20\%.
Notably, for the CS dataset, GNNGuard’s performance decreases are marginal. This discrepancy arises because more than 80\% of poisoning edges in Pubmed are merged into supernodes during coarsening, while less than 30\% are merged in the CS dataset.

Our results suggest that, defensive models that rely on identifying and eliminating suspicious edges may not perform well with graph coarsening based reduction. Sparsification methods are generally safer to use with defensive GNNs, but graph coarsening methods require particular caution. While coarsening can accelerate computation, it risks propagating adversarial effects by merging poisoned edges into supernodes. This can undermine even robust GNNs, as the inherited perturbations within supernodes are difficult to detect and mitigate.

\subsection{Evaluating Graph Reduction under Evasion Attacks}
Besides GMAs, we also implement two evasion attacks: a feature-based modification attack, namely, InfMax \cite{ma2022infmax}, and a graph injection attack, namely, AGIA \cite{chen2022agia}.
Although these attacks modify the graph data during the inference time and do not affect the GNN training accelerated under graph reduction, we are interested in how the evasion attacks designed for original graphs are affected when the model is trained on reduced clean graphs produced by reduction methods. The results are presented in Tables \ref{tab:other_acc_spar}.

It is noteworthy that, unlike GMAs, which set the perturbation ratio by counting modified edges, feature-based and injection attacks use the number of modified/injected nodes for their perturbation ratios, which is not comparable to the perturbation ratio used in our previous experiments. Therefore, we follow the setting from their original papers to examine the interaction between these attacks and graph reduction methods. The results show that $ACC_r \approx ACC_p$ across every reduction method. This indicates that accelerated training using graph reduction does not affect the model’s vulnerability to evasion attacks. The model trained on reduced graphs retains similar robustness characteristics to the model trained on the original graph when evaluated under evasion settings.
\begin{table*}[ht!]
\centering
\caption{GCN's poisoned accuracy under $p=0.05\%$, and post-reduction accuracy after graph sparsification.}
\begin{tabular}{cc|c|cccc|cccccc}
\toprule
\multicolumn{2}{c|}{}  & \multicolumn{1}{c|}{} & \multicolumn{10}{c}{\centering $ACC_r$}\\
Dataset & Attack & $ACC_p$ & RNE & LD & LS & SCAN & VN & VE & VC & HE & JC & KRON\\
\hline
\multirow{2}{*}{Cora}
& AGIA & 70.99\% &
    70.67\% & 71.56\% & 68.15\% & 69.86\% &
    76.18\% & 75.10\% & 74.88\% & 74.71\% & 75.56\% & 75.39\%\\
& InfMax & 76.90\% & 
    76.20\% & 77.65\% & 75.16\% & 75.28\% &
    80.47\% & 78.72\% & 76.97\% & 79.08\% & 80.05\% & 80.69\%\\
\hline
\multirow{2}{*}{Pubmed}
& AGIA & 79.18\% &
    80.25\% & 79.08\% & 79.57\% & 80.00\% &
    77.98\% & 77.12\% & 77.59\% & 77.45\% & 77.99\% & 78.47\%\\
& InfMax & 62.30\% &
    61.90\% & 63.72\% & 62.30\% & 63.29\% &
    68.17\% & 66.44\% & 66.69\% & 66.65\% & 66.97\% & 67.04\%\\
\hline
\multirow{2}{*}{CS}
& AGIA & 92.11\% &
    91.95\% & 92.04\% & 91.77\% & 91.65\% &
    90.85\% & 90.47\% & 90.85\% & 91.15\% & 90.86\% & 91.04\%\\
& InfMax & 90.86\% & 
    91.29\% & 91.07\% & 91.13\% & 91.20\% &
    88.12\% & 88.97\% & 88.76\% & 89.29\% & 89.79\% & 89.56\%\\
\hline
\end{tabular}
\label{tab:other_acc_spar}
\end{table*}

\section{Conclusion}
This paper explores the impact of graph reduction techniques—sparsification and coarsening—on the robustness of Graph Neural Networks (GNNs) under adversarial attacks. 
Our findings highlight the contrasting effects of sparsification and coarsening. Graph sparsification effectively removes poisoned edges during training, mitigating the impact of certain poisoning attacks, such as Mettack. However, it shows limited effectiveness against evasion attacks, like PGD, which target the inference phase. On the other hand, graph coarsening amplifies adversarial vulnerabilities by merging poisoned edges into supernodes, creating noisy and unclean graph representations that degrade model performance, even for robust GNNs.
Additionally, we show that combining robust GNNs with graph sparsification preserves or enhances their defense capabilities. Conversely, coarsening disrupts the effectiveness of these defensive models by transferring adversarial effects to supernodes, undermining their ability to eliminate perturbations. These insights emphasize the need for careful selection and evaluation of graph reduction methods when designing scalable and robust GNN systems.

Overall, our findings offer practical insights for building scalable and robust GNN systems. Among the graph reduction techniques evaluated, sparsification—particularly the Local Degree algorithm—emerges as a strong candidate for deployment in security-critical applications due to its higher robustness against poisoning attacks. In contrast, graph coarsening tends to amplify adversarial effects, which are insufficiently mitigated by existing defenses. We therefore advise caution when applying graph coarsening for GNN training acceleration in security-sensitive scenarios. The reduction ratio is another critical factor influencing GNN robustness. Our results indicate a significant drop in accuracy when the reduction ratio falls below 0.3, a trend observed across both sparsification and coarsening methods. Thus, overly aggressive reductions should be avoided to preserve model robustness.
Future work could extend this investigation by exploring hybrid graph reduction approaches or developing new reduction methods that explicitly account for adversarial robustness.

{\small
\bibliographystyle{splncs04}
\bibliography{main}
}

\clearpage
\appendix
\section{Supplementary Appendix}
\subsection{Descriptions of Graph Reduction Methods}
\label{sec:Appendix_A.1}
\noindent\textbf{Graph Coarsening.}~ Six methods for graph coarsening are listed as follows:
\begin{itemize}
    \item Variation Neighbourhoods ($VN$)), Variation Cliques ($VC$), and Variation Edges ($VE$) are methods based on spectral principles, as outlined by Loukas (2019). The process begins by computing the graph Laplacian matrix $\mathbf{L}$. With a target graph dimension $n$, the aim is to obtain a coarsened Laplacian matrix $\mathbf{L}_c$ of size $n \times n$ that closely approximates $\mathbf{L}$ while keeping the graph information loss $\epsilon$ below a predefined threshold. The primary difference among the $VN$, $VC$, and $VE$ methods lies in their selection of node sets for contraction. In the $VN$ approach, each vertex along with its neighboring vertices forms a candidate set. For $VC$, all maximal cliques identified using the Bron–Kerbosch algorithm serve as individual candidate sets. In the $VE$ method, each edge is treated as a separate candidate set. These candidate sets are sorted based on cost, and a recursive computation is performed: the set with the lowest cost is selected first, and the nodes within this set are coarsened. This process repeats until the updated $\mathbf{L}_c$ reaches the desired size.
    \item Heavy Edge Matching(HE): In the Heavy Edge Matching approach, edge pairs are selected for contraction at each coarsening level by computing the Maximum Weight Matching, where the weight of an edge pair is determined based on the maximum vertex degree within the pair \cite{loukas2018spectrally}. This strategy tends to contract edges that are peripheral to the main structure of the graph, thereby preserving its core integrity.

    \item Algebraic JC(JC): This method calculates algebraic distances as weights for each candidate set of edges. The distances are computed from test vectors, each obtained through iterations of the Jacobi relaxation algorithm \cite{ron2011relaxation}.

    \item Kron Reduction(KRON): At each coarsening stage of the Kron Reduction method \cite{dorfler2012kron}, a subset of vertices is selected based on the positive entries of the final eigenvector of the Laplacian matrix. The graph's size is then reduced through Kron Reduction, aiming to preserve its spectral characteristics for efficient analysis of large-scale networks.
\end{itemize}

\noindent\textbf{Graph Sparsification.}~ Four methods for graph sparsification are listed as follows:
\begin{itemize}
    \item Random Node Edge (RNE): This method uniformly and randomly selects both nodes and edges to retain in the sparsified graph.

    \item Local Degree(LD) \cite{hamann2016structure}: Each node ranks its neighboring nodes based on their degrees and selects a fraction of the top-ranked neighbors. This approach ensures that every node retains at least one edge.
    
    \item Local Similarity(LS) \cite{satuluri2011local}: This technique calculates the Jaccard similarity scores between each vertex and its neighbors. Edges with the highest local similarity scores are then selected for inclusion in the sparsified graph.
    
    \item SCAN \cite{xu2007scan}: SCAN computes similarity scores for all pairs of vertices in the graph using the SCAN similarity measure. It sorts the edges based on these scores and includes those with the highest similarity in the sparsified graph.
\end{itemize}

\begin{table}
\centering
\caption{Statistics for each dataset}
\begin{tabular}{c|rrrr} 
Dataset & \#Nodes & \#Edges & \#Featuress & \#Classes\\
\hline
Cora & 2703 & 10556 & 1433 & 7\\
Pubmed & 19717 & 88648 & 500 & 3\\
CS & 18333 & 168788 & 6805 & 15\\
\hline
\end{tabular}
\label{tab:datasets}
\end{table}
\begin{table}
\centering
\scriptsize
\caption{Label difference ratio between original node's label and its corresponding super node's label by VC coarsening under Cora and Pubmed datasets with $r=0.3$ against five poisoning attacks}
\begin{tabular}{cr|rr} 
\toprule
Attack & Dataset & Clean Label Diff Ratio & Poisoned Label Diff Ratio\\
\hline
\multirow{3}{*}{DICE}
& Cora & 40.60\% & 41.81\%\\
& Pubmed & 41.16\% & 44.64\%\\
& CS & 45.66\% & 47.89\%\\
\hline
\multirow{3}{*}{NEA}
& Cora & 41.85\% & 44.14\%\\
& Pubmed & 43.08\% & 42.30\%\\
& CS & 45.21\% & 49.22\%\\
\hline
\multirow{3}{*}{PGD}
& Cora & 42.95\% & 44.83\%\\
& Pubmed & 42.75\% & 41.25\%\\
& CS & 45.44\% & 47.08\%\\
\hline
\multirow{3}{*}{Mettack}
& Cora & 40.72\% & 42.37\%\\
& Pubmed & 42.53\% & 53.02\%\\
& CS & 45.21\% & 49.47\%\\
\hline
\multirow{3}{*}{PRBCD}
& Cora & 43.34\% & 46.88\%\\
& Pubmed & 41.72\% & 41.26\%\\
& CS & 45.21\% & 48.90\%\\
\hline
\end{tabular}
\label{tab:complete_label_diff}
\end{table}

\begin{figure*}
\centering
  \includegraphics[width=0.75\textwidth]{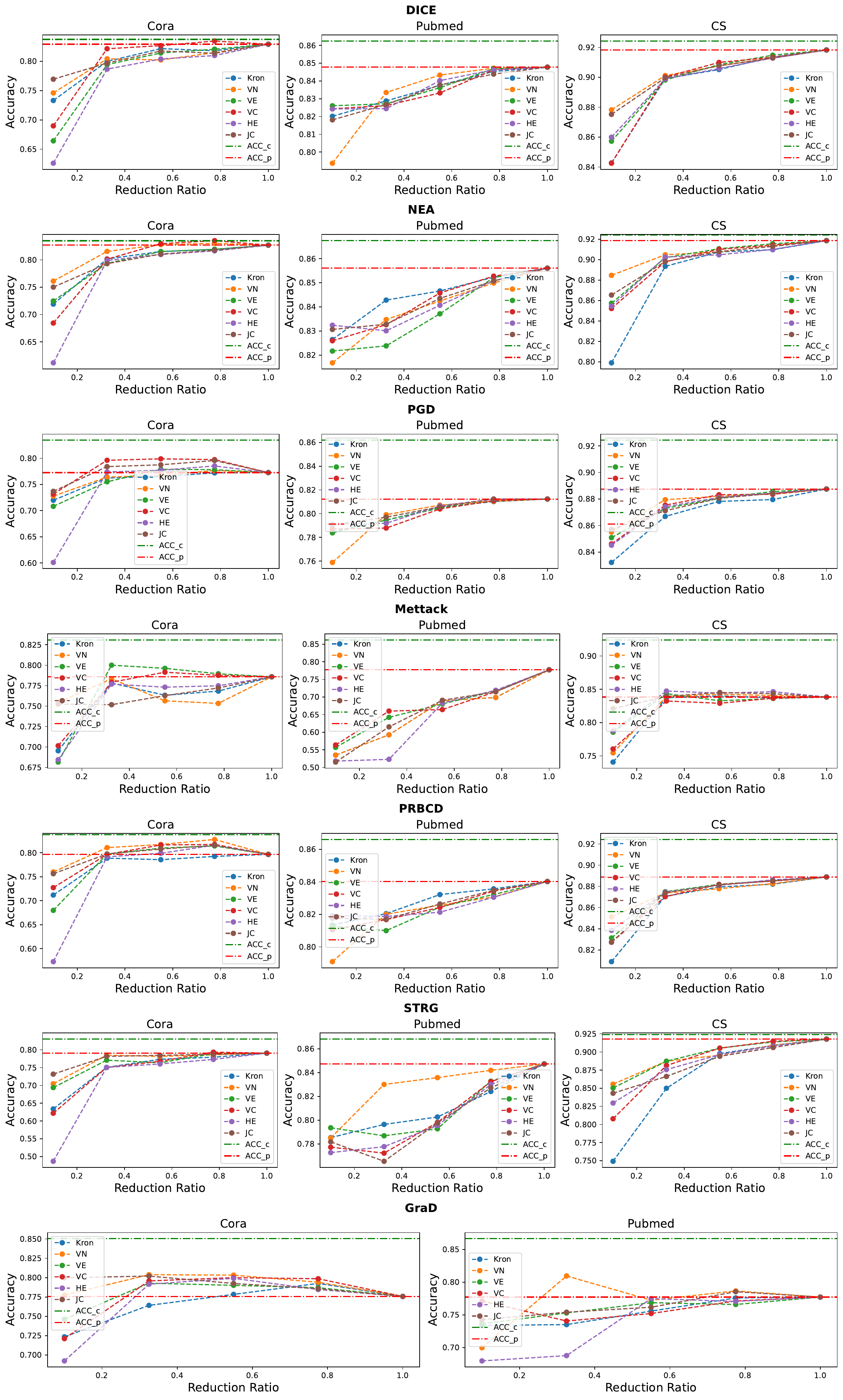}
  \caption{$ACC_r$ using six coarsening algorithms against seven attacks with various $r$ in three datasets. Overall, $ACC_r$ sharply decreases as the reduction ratio is less than 0.3.}
  \label{fig:complete_coarsening_gcn}
\end{figure*}

\begin{figure*}
\centering
  \includegraphics[width=0.75\textwidth]{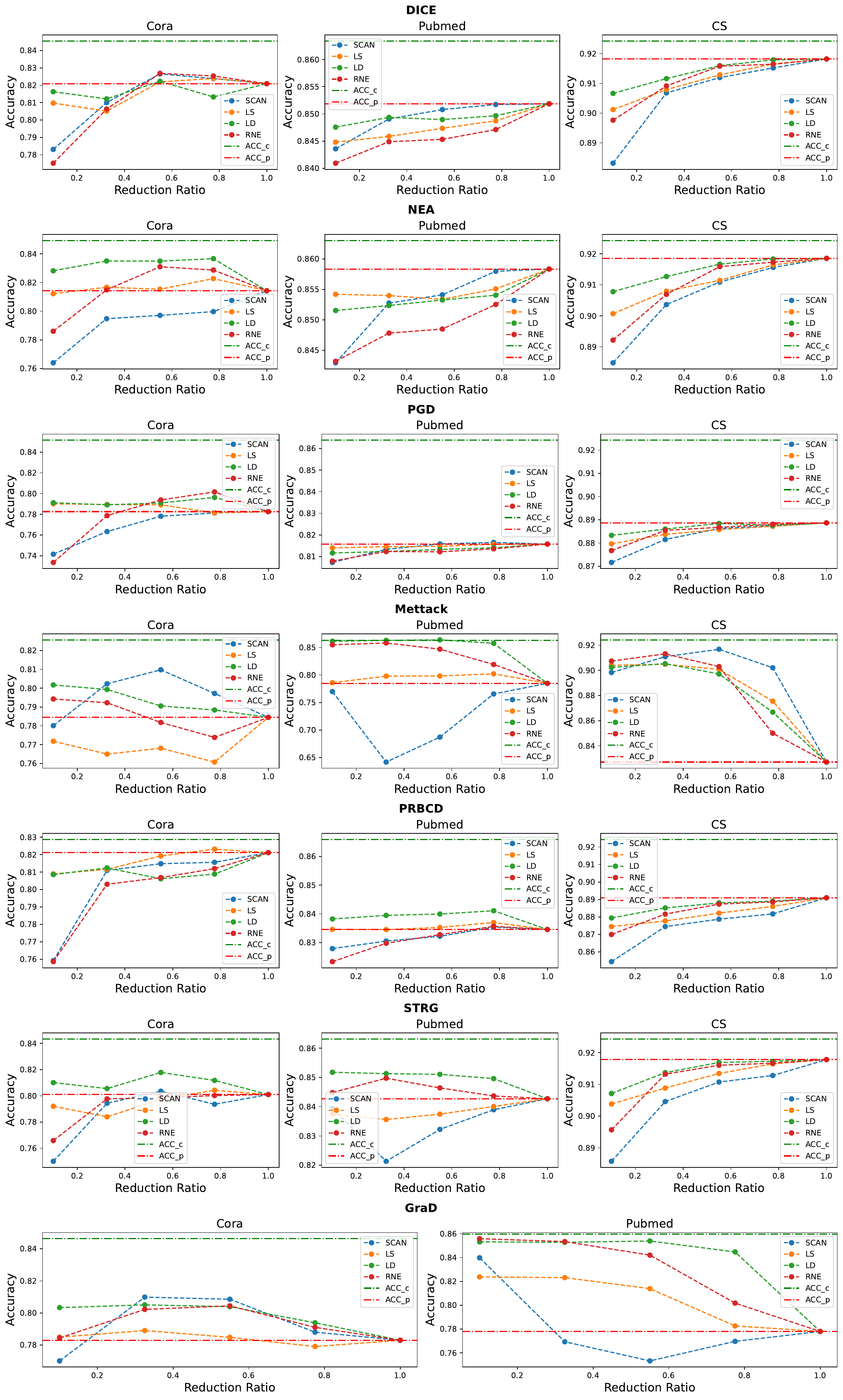}
  \caption{$ACC_r$ using four sparsification algorithms against seven attacks with various $r$ in three datasets. Mettack and GraD are notably mitigated, as evidenced by increased $ACC_r$ with decreasing $r$, while other attacks exhibit relatively stable $ACC_r$ across different reduction ratios.}
  \label{fig:complete_sparsification_gcn}
\end{figure*}

\begin{figure*}
\centering
  \includegraphics[width=0.75\textwidth]{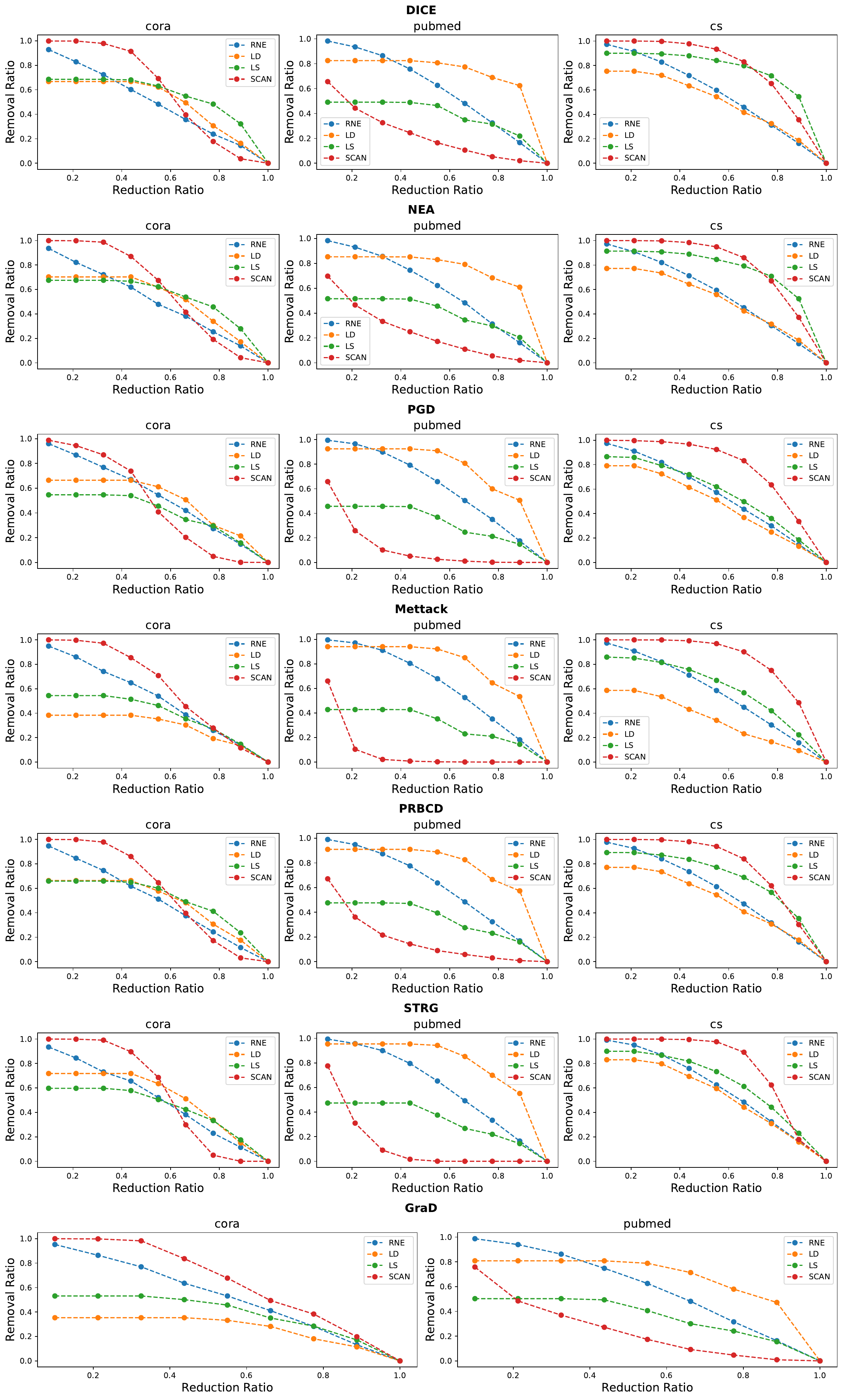}
  \caption{Complete elimination ratios of newly-added perturbation edges by graph sparsification}
  \label{fig:complete_merge_compare}
\end{figure*}

\begin{figure*}
\centering
  \includegraphics[width=0.75\textwidth]{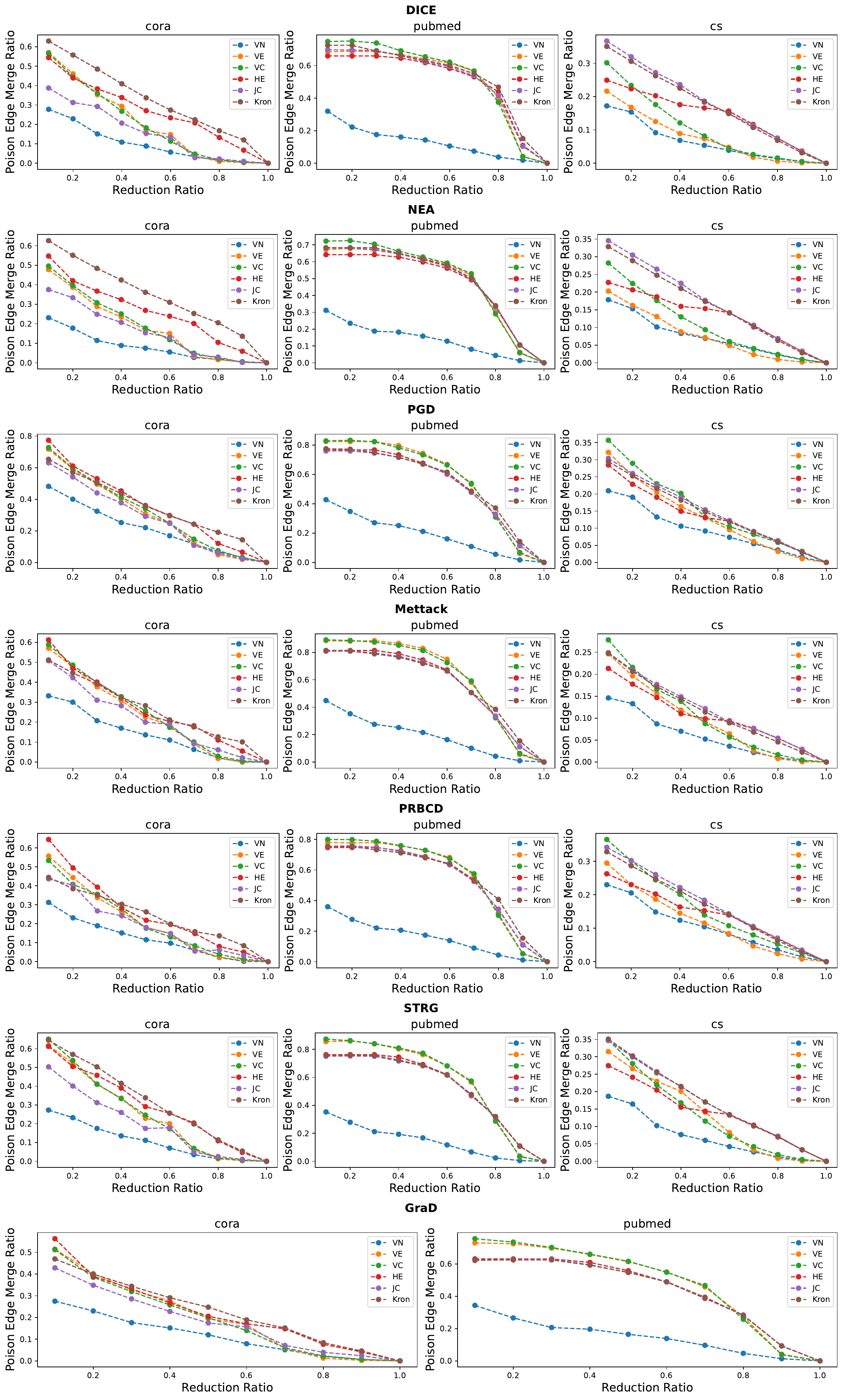}
  \caption{Complete merge ratios of malicious perturbation edges under graph coarsening}
  \label{fig:complete_coarsen_edge_merge}
\end{figure*}

\begin{figure*}
  \includegraphics[width=1\textwidth]{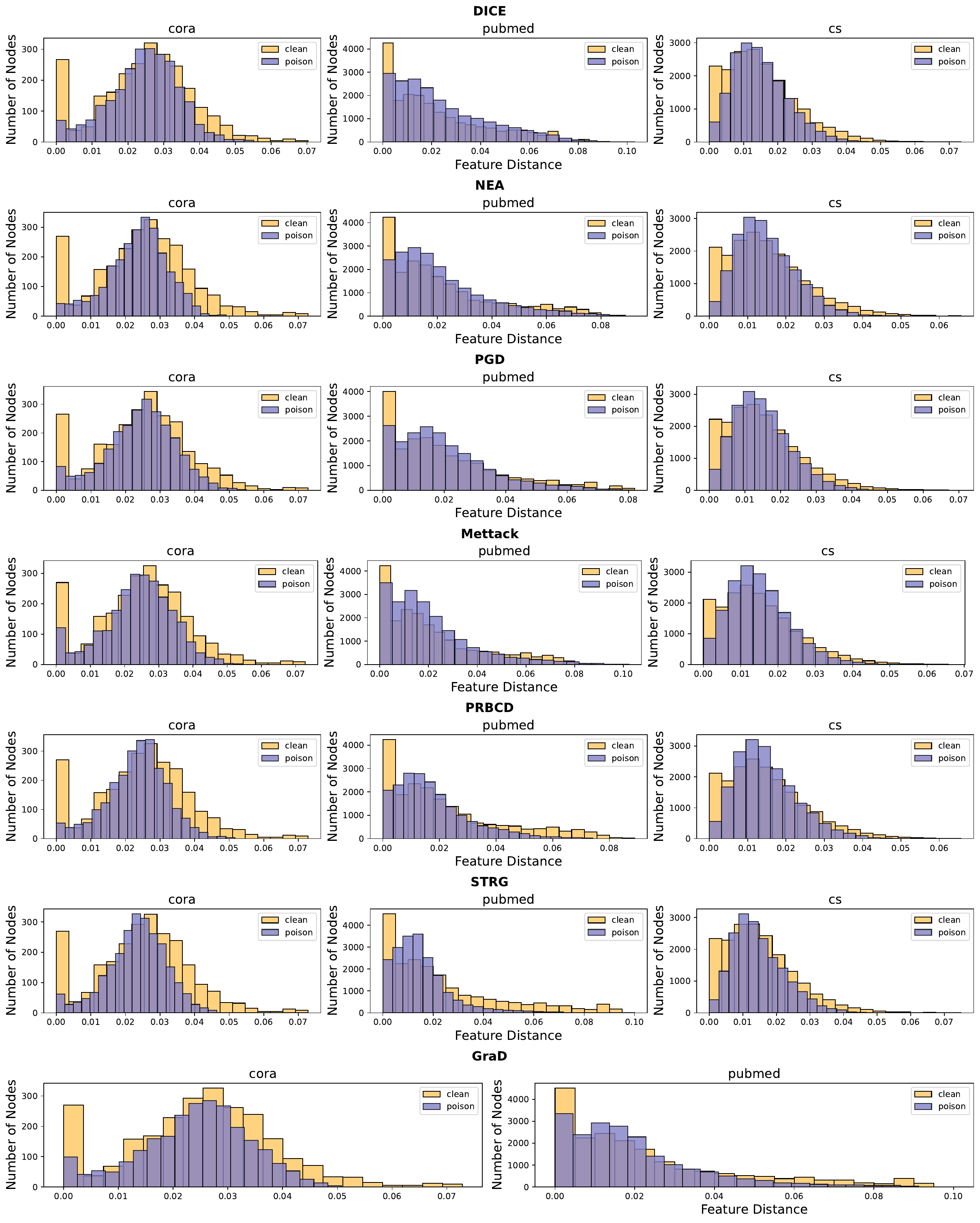}
  \caption{Complete feature distance between clean coarsened graph and poisoned coarsened graph}
  \label{fig:complete_feature_distance_compare}
\end{figure*}

\begin{table*}[ht]
\footnotesize
\centering
\caption{Cora's clean accuracy, poisoned accuracy under $p=0.05\%$, and reduced accuracy after graph sparsification under $r=0.325$ with various GNN architecture}
\setlength{\tabcolsep}{2pt} 
\begin{tabular}{cr|rr|rrrr|rrrrrr}
\toprule
\multicolumn{1}{l}{}  & \multicolumn{1}{l|}{} & \multicolumn{1}{l}{} & \multicolumn{1}{l|}{} & \multicolumn{10}{c}{$ACC_r$}\\
\multicolumn{1}{l}{Attack}  & \multicolumn{1}{l|}{GNN} & \multicolumn{1}{c}{$ACC_c$} & \multicolumn{1}{l|}{$ACC_p$} & RNE & LD & LS & SCAN & VN & VE & VC & HE & JC & KRON\\
\hline
\multirow{6}{*}{DICE}
& GCN & 84.06\% & 81.98\%
    & 80.63\% & 81.22\% & 80.51\% & 81.01\% & 
    80.42\% & 79.44\% & 82.14\% & 78.66\% & 79.7\% & 79.98\%\\
& GAT & 84.55\% & 81.41\%
    & 81.26\% & 81.29\% & 81.83\% & 81.14\% & 
    80.01\% & 79.86\% & 81.18\% & 78.87\% & 79.72\% & 81.38\%\\
& SAGE & 83.63\% & 82.01\%
    & 76.33\% & 76.63\% & 79.31\% & 77.06\% & 
    80.02\% & 79.68\% & 80.01\% & 77.22\% & 79.92\% & 79.41\%\\
& GNNGuard & 80.59\% & 77.59\%
    & 73.60\% & 75.58\% & 75.82\% & 74.79\% & 
    \textbf{79.9\%} & 77.99\% & 78.62\% & 76.4\% & 78.42\% & 77.84\%\\
& RGCN & 83.82\% & 81.73\%
    & 80.60\% & 81.04\% & 81.33\% & 80.56\% & 
    N/A & N/A & N/A & N/A & N/A & N/A\\
& Median & 84.52\% & 83.15\%
    & 80.79\% & 81.65\% & 81.21\% & 80.92\% & 
    79.87\% & 80.89\% & 81.71\% & 79.14\% & 81.24\% & 80.3\%\\
\hline
\multirow{6}{*}{NEA}
& GCN & 84.06\% & 82.46\%
    & 81.49\% & \textbf{83.50\%} & 81.67\% & 79.48\% & 
    81.55\% & 79.37\% & 80.12\% & 79.89\% & 79.31\% & 80.17\%\\
& GAT & 84.55\% & 82.70\%
    & 82.57\% & 80.54\% & 81.83\% & 80.48\% & 
    80.21\% & 80.54\% & 79.56\% & 79.99\% & 80.68\% & 81.08\%\\
& SAGE & 83.63\% & 81.79\%
    & 75.7\% & 77.48\% & 79.33\% & 76.85\% & 
    80.4\% & 79.34\% & 78.71\% & 78.81\% & 77.88\% & 77.65\%\\
& GNNGuard & 80.59\% & 79.06\%
    & 75.7\% & 77.48\% & 79.33\% & 76.85\% & 
    \textbf{80.27\%} & 79.1\% & \textbf{79.84\%} & 77.59\% & 77.09\% & 77.11\%\\
& RGCN & 83.82\% & 82.34\%
    & 81.15\% & 80.81\% & 82.74\% & 79.35\% & 
    N/A & N/A & N/A & N/A & N/A & N/A\\
& Median & 84.52\% & \textbf{83.88\%}
    & 82.01\% & 80.34\% & 82.38\% & 78.41\% & 
    81.48\% & 80.29\% & 81.86\% & 80.08\% & 80.35\% & 79.86\%\\
\hline
\multirow{6}{*}{PGD}
& GCN & 84.06\% & 78.20\%
    & 77.88\% & 78.94\% & 78.91\% & 76.34\% & 
    76.41\% & 75.51\% & 79.62\% & 77.37\% & 78.4\% & 76.24\%\\
& GAT & 84.55\% & 78.25\%
    & 78.43\% & 78.12\% & 79.14\% & 77.6\% & 
    76.82\% & 76.4\% & 79.58\% & 77.08\% & 77.72\% & 76.07\%\\
& SAGE & 83.63\% & 78.52\%
    & 74.64\% & 76.44\% & 75.59\% & 74.65\% & 
    76.68\% & 76.03\% & 78.53\% & 75.39\% & 77.79\% & 75.44\%\\
& GNNGuard & 80.59\% & 77.04\%
    & 73.27\% & 75.94\% & 76.69\% & 71.84\% & 
    77.43\% & 76.39\% & 78.35\% & 76.44\% & 78.2\% & 75.66\%\\
& RGCN & 83.82\% & 78.23\%
    & 78.01\% & 78.21\% & 79.08\% & 76.57\% & 
    N/A & N/A & N/A & N/A & N/A & N/A\\
& Median & 84.52\% & 79.14\%
    & 78.02\% & 78.51\% & 78.49\% & 77.59\% & 
    78.08\% & 76.59\% & 78.74\% & 78.01\% & 79.0\% & 76.48\%\\
\hline
\multirow{6}{*}{Mettack}
& GCN & 84.06\% & 74.46\%
    & 79.23\% & 76.5\% & 79.93\% & 80.23\% & 
    78.34\% & 80.0\% & 77.92\% & 77.7\% & 75.17\% & 77.82\%\\
& GAT & 84.55\% & 77.83\%
    & 79.96\% & 77.98\% & 81.9\% & 80.89\% & 
    78.1\% & 79.76\% & 79.7\% & 79.47\% & 79.97\% & 77.27\%\\
& SAGE & 83.63\% & 76.68\%
    & 73.99\% & 74.07\% & 77.97\% & 76.96\% & 
    77.47\% & 78.88\% & 76.84\% & 77.51\% & 74.98\% & 75.53\%\\
& GNNGuard & 80.59\% & 76.30\%
    & 72.95\% & 71.92\% & 75.93\% & 74.55\% & 
    76.47\% & 76.54\% & 74.24\% & 74.89\% & 74.35\% & 75.04\%\\
& RGCN & 83.82\% & 73.75\%
    & 77.83\% & 75.92\% & 78.34\% & 80.27\% & 
    N/A & N/A & N/A & N/A & N/A & N/A\\
& Median & 84.52\% & 77.64\%
    & 79.22\% & 77.2\% & 81.91\% & 79.98\% & 
    78.27\% & 80.18\% & 79.54\% & 78.57\% & 79.42\% & 77.7\%\\
\hline
\multirow{6}{*}{PRBCD}
& GCN & 84.06\% & 81.11\%
    & 80.3\% & 81.16\% & 81.24\% & 81.1\% & 
    81.09\% & 79.74\% & 79.69\% & 79.16\% & 79.76\% & 78.88\%\\
& GAT & 84.55\% & 81.84\%
    & 79.50\% & 81.78\% & 79.94\% & 80.31\% & 
    79.81\% & 80.11\% & 79.08\% & 79.57\% & 81.24\% & 78.18\%\\
& SAGE & 83.63\% & 80.78\%
    & 77.16\% & 79.16\% & 75.7\% & 76.6\% & 
    79.98\% & 78.06\% & 79.89\% & 76.66\% & 80.52\% & 76.32\%\\
& GNNGuard & 80.59\% & 77.11\%
    & 74.74\% & 77.01\% & 77.03\% & 74.79\% & 
    79.31\% & 79.28\% & 78.56\% & 77.77\% & 78.62\% & 75.35\%\\
& RGCN & 83.82\% & 80.46\%
    & 79.59\% & 80.31\% & 80.2\% & 80.35\% & 
    N/A & N/A & N/A & N/A & N/A & N/A\\
& Median & 84.52\% & 81.57\%
    & 79.52\% & 80.99\% & 79.87\% & 79.15\% & 
    81.23\% & 81.28\% & 79.96\% & 79.57\% & 80.68\% & 80.06\%\\
\hline
\multirow{6}{*}{STRG-Heuristic}
& GCN & 84.06\% & 79.07\% &
79.77\% & 80.55\% & 78.40\% & 79.44\% & 
78.51\% & 77.09\% & 75.04\% & 75.07\% & 78.17\% & 75.07\%\\
& GAT & 84.55\% & 80.84\% &
78.49\% & 80.96\% & 79.13\% & 79.52\% &
77.95\% & 78.87\% & 77.59\% & 78.42\% & 80.78\% & 77.90\%\\
& SAGE & 83.63\% & 79.32\% &
74.04\% & 77.13\% & 74.09\% & 75.20\% &
76.25\% & 73.15\% & 72.44\% & 74.60\% & 74.00\% & 74.08\%\\
& GNNGuard & 80.59\% & 76.71\% &
73.68\% & 75.41\% & 74.07\% & 74.52\% & 
75.03\% & 75.05\% & 72.62\% & 73.84\% & 74.29\% & 72.27\%\\
& RGCN & 83.82\% & 77.89\% &
78.64\% & 80.01\% & 77.09\% & 79.16\% & 
N/A & N/A & N/A & N/A & N/A & N/A\\
& Median & 84.52\% & 80.89\% & 
78.92\% & 79.76\% & 77.68\% & 77.78\% & 
78.15\% & 77.61\% & 77.38\% & 76.25\% & 76.77\% & 76.36\%\\
\hline
\multirow{6}{*}{GraD}
& GCN & 84.06\% & 76.26\% &
80.22\% & 80.50\% & 78.90\% & 80.99\% & 
80.38\% & 79.29\% & 79.58\% & 79.15\% & 80.20\% & 76.42\%\\
& GAT & 84.55\% & 80.86\% &
81.44\% & 82.25\% & 79.21\% & 80.54\% & 
81.47\% & 78.85\% & 79.66\% & 79.75\% & 80.51\% & 81.47\%\\
& SAGE & 83.63\% & 79.53\% & 
76.91\% & 78.43\% & 77.39\% & 77.22\% & 
78.41\% & 76.79\% & 76.59\% & 77.64\% & 77.96\% & 75.37\%\\
& GNNGuard & 80.59\% & 78.40\% &
74.56\% & 76.22\% & 74.92\% & 75.39\% & 
76.98\% & 76.00\% & 76.30\% & 75.01\% & 76.90\% & 73.31\%\\
& RGCN & 83.82\% & 74.30\% &
79.38\% & 78.87\% & 77.40\% & 79.95\% &
N/A & N/A & N/A & N/A & N/A & N/A\\
& Median & 84.52\% & 81.30\% &
81.74\% & 82.04\% & 79.33\% & 78.91\% & 
81.84\% & 79.72\% & 80.78\% & 79.47\% & 81.37\% & 81.55\%\\
\hline
\label{tab:complete_cora_gnns_reduced_acc}
\end{tabular}
\end{table*}

\begin{table*}[ht]
\footnotesize
\centering
\caption{Pubmed's clean accuracy, poisoned accuracy under $p=0.05\%$, and reduced accuracy after graph sparsification under $r=0.325$ with various GNN architecture}
\setlength{\tabcolsep}{2pt} 
\begin{tabular}{cr|rr|rrrr|rrrrrr}
\toprule
\multicolumn{1}{l}{}  & \multicolumn{1}{l|}{} & \multicolumn{1}{l}{} & \multicolumn{1}{l|}{} & \multicolumn{10}{c}{$ACC_r$}\\
\multicolumn{1}{l}{Attack}  & \multicolumn{1}{l|}{GNN} & $ACC_c$ & \multicolumn{1}{l|}{$ACC_p$} & RNE & LD & LS & SCAN & VN & VE & VC & HE & JC & KRON\\
\hline
\multirow{6}{*}{DICE}
& GCN & 86.15\% & \textbf{85.21\%} 
    & 84.49\% & 84.94\% & 84.59\% & 84.36\% & 
    83.35\% & 82.72\% & 82.62\% & 82.44\% & 82.67\% & 82.87\%\\
& GAT & 85.13\% & 83.83\% 
    & 82.59\% & 83.29\% & 83.00\% & 83.14\% & 
    82.80\% & 81.89\% & 80.72\% & 81.53\% & 79.53\% & 80.72\%\\
& SAGE & 85.59\% & \textbf{85.10\%} 
    & \textbf{85.60\%} & \textbf{85.06\%} & \textbf{85.40\%} & \textbf{85.88\%} & 
    82.12\% & 77.36\% & 79.01\% & 78.24\% & 79.57\% & 78.65\%\\
& GNNGuard & 84.67\% & \textbf{84.52\%}
    & \textbf{84.12\%} & \textbf{84.40\%} & \textbf{84.20\%} & \textbf{84.54\%} & 
    83.41\% & 81.85\% & 82.15\% & 82.14\% & 82.05\% & 82.21\%\\
& RGCN & 85.49\% & \textbf{84.63\%} 
    & 83.17\% & 84.22\% & 84.11\% & 84.35\% & 
    N/A & N/A & N/A & N/A & N/A & N/A\\
& Median & 84.43\% & 83.15\% 
    & 82.86\% & \textbf{83.47\%} & \textbf{83.46\%} & 83.42\% & 
    83.31\% & 80.41\% & 81.56\% & 80.96\% & 81.27\% & 82.21\%\\
\hline
\multirow{6}{*}{NEA}
& GCN & 86.15\% & \textbf{85.49\%} 
    & 84.78\% & \textbf{85.24\%} & \textbf{85.40\%} & \textbf{85.28\%} & 
    83.47\% & 82.38\% & 83.26\% & 83.01\% & 83.28\% & 84.28\%\\
& GAT & 85.13\% & \textbf{84.50\%} 
    & 83.24\% & 83.79\% & 83.41\% & 83.46\% & 
    82.78\% & 80.66\% & 80.18\% & 82.07\% & 80.47\% & 81.78\%\\
& SAGE & 85.59\% & \textbf{85.69\%} 
    & \textbf{85.76\%} & \textbf{85.81\%} & \textbf{85.74\%} & \textbf{85.62\%} & 
    82.28\% & 78.43\% & 78.47\% & 77.70\% & 78.89\% & 78.76\%\\
& GNNGuard & 84.67\% & \textbf{84.70\%} 
    & \textbf{84.58\%} & \textbf{84.25\%} & \textbf{84.66\%} & \textbf{84.85\%} & 
    83.21\% & 82.45\% & 83.23\% & 82.81\% & 82.50\% & 83.47\%\\
& RGCN & 85.49\% & \textbf{85.17\%} 
    & \textbf{84.13\%} & \textbf{84.78\%} & \textbf{84.54\%} & \textbf{84.58\%} & 
    N/A & N/A & N/A & N/A & N/A & N/A\\
& Median & 84.43\% & \textbf{84.19\%} 
    & \textbf{83.60\%} & \textbf{83.84\%} & \textbf{83.84\%} & \textbf{84.16\%} & 
    83.61\% & 81.78\% & 81.94\% & 81.67\% & 81.51\% & 82.60\%\\
\hline
\multirow{6}{*}{PGD}
& GCN & 86.15\% & 81.33\% 
    & 81.23\% & 81.17\% & 81.46\% & 81.34\% & 
    79.91\% & 79.46\% & 78.78\% & 79.19\% & 79.70\% & 79.45\%\\
& GAT & 85.13\% & 80.67\% 
    & 80.36\% & 80.78\% & 80.51\% & 80.11\% & 
    79.61\% & 73.47\% & 74.97\% & 78.51\% & 78.25\% & 78.84\%\\
& SAGE & 85.59\% & 83.35\% 
    & 83.89\% & 82.86\% & 83.74\% & \textbf{84.67\%} & 
    79.48\% & 75.09\% & 75.32\% & 77.47\% & 77.82\% & 76.91\%\\
& GNNGuard & 84.67\% & 83.53\% 
    & 83.63\% & 83.53\% & \textbf{83.70\%} & \textbf{84.19\%} & 
    82.57\% & 81.32\% & 81.06\% & 81.17\% & 81.52\% & 81.80\%\\
& RGCN & 85.49\% & 80.70\% 
    & 80.48\% & 80.77\% & 80.74\% & 80.46\% & 
    N/A & N/A & N/A & N/A & N/A & N/A\\
& Median & 84.43\% & 80.82\% 
    & 80.62\% & 81.12\% & 80.70\% & 80.94\% & 
    80.14\% & 78.22\% & 78.64\% & 78.45\% & 79.18\% & 79.70\%\\
\hline
\multirow{6}{*}{Mettack}
& GCN & 86.15\% & 78.07\% 
    & \textbf{85.80\%} & \textbf{86.27\%} & 79.78\% & 64.18\% & 
    64.20\% & 59.41\% & 61.76\% & 45.27\% & 61.97\% & 64.21\%\\
& GAT & 85.13\% & 82.95\% 
    & 83.93\% & \textbf{84.47\%} & 79.66\% & 64.84\% & 
    64.20\% & 59.41\% & 61.76\% & 45.27\% & 61.97\% & 64.21\%\\
& SAGE & 85.59\% & 81.40\% 
    & \textbf{85.36\%} & \textbf{86.21\%} & 83.28\% & 74.56\% & 
    62.39\% & 54.26\% & 55.58\% & 45.41\% & 54.83\% & 62.39\%\\
& GNNGuard & 84.67\% & \textbf{84.67\%} 
    & \textbf{84.54\%} & \textbf{84.80\%} & \textbf{84.60\%} & \textbf{84.65\%} & 
    63.64\% & 61.48\% & 60.69\% & 48.01\% & 62.26\% & 63.64\%\\
& RGCN & 85.49\% & 81.80\% 
    & \textbf{85.19\%} & \textbf{85.17\%} & 81.28\% & 80.67\% & 
    N/A & N/A & N/A & N/A & N/A & N/A\\
& Median & 84.43\% & 79.39\% 
    & \textbf{83.97\%} & \textbf{84.12\%} & 78.09\% & 67.04\% & 
    62.01\% & 57.65\% & 61.41\% & 46.91\% & 60.34\% & 62.01\%\\
\hline
\multirow{6}{*}{PRBCD}
& GCN & 86.15\% & 83.39\% 
    & 82.97\% & 83.95\% & 83.45\% & 83.05\% & 
    82.01\% & 81.00\% & 81.68\% & 81.87\% & 81.70\% & 82.04\%\\
& GAT & 85.13\% & 82.56\% 
    & 81.88\% & 82.72\% & 82.40\% & 81.69\% & 
    80.71\% & 77.03\% & 76.78\% & 80.19\% & 79.72\% & 81.45\%\\
& SAGE & 85.59\% & \textbf{85.08\%} 
    & \textbf{84.67\%} & \textbf{84.85\%} & \textbf{85.31\%} & \textbf{85.53\%} & 
     81.50\% & 72.98\% & 77.33\% & 73.47\% & 79.11\% & 79.75\%\\
& GNNGuard & 84.67\% & \textbf{84.20\%} 
    & \textbf{84.03\%} & \textbf{84.34\%} & \textbf{84.12\%} & \textbf{84.14\%} & 
    83.61\% & 81.87\% & 82.28\% & 82.68\% & 82.76\% & 82.56\%\\
& RGCN & 85.49\% & 83.18\% 
    & 82.57\% & 83.56\% & 83.10\% & 82.53\% & 
    N/A & N/A & N/A & N/A & N/A & N/A\\
& Median & 84.43\% & 83.33\% 
    & 82.46\% & 83.40\% & 82.79\% & 82.60\% & 
    82.16\% & 80.17\% & 80.89\% & 78.98\% & 81.50\% & 81.59\%\\
\hline
\multirow{6}{*}{STRG-Heuristic}
& GCN & 86.15\% & 84.73\% & 
84.98\% & 85.13\% & 83.56\% & 85.13\% & 
83.01\% & 78.69\% & 77.23\% & 77.78\% & 76.54\% & 79.64\%\\
& GAT & 85.13\% & 83.81\% & 
83.00\% & 83.48\% & 82.32\% & 81.27\% &
82.18\% & 74.13\% & 77.71\% & 71.20\% & 77.14\% & 76.30\%\\
& SAGE & 85.59\% & 85.67\% & 
85.33\% & 85.11\% & 85.28\% & 82.35\% & 
82.41\% & 67.51\% & 74.68\% & 64.31\% & 78.18\% & 79.74\%\\
& GNNGuard & 84.67\% & 84.30\% & 
84.29\% & 84.40\% & 83.91\% & 83.76\% & 
82.88\% & 79.71\% & 79.17\% & 77.93\% & 79.07\% & 80.93\%\\
& RGCN & 85.49\% & 84.46\% & 
84.16\% & 84.50\% & 83.47\% & 82.53\% & 
N/A & N/A & N/A & N/A & N/A & N/A\\
& Median & 84.43\% & 82.60\% &
83.05\% & 83.43\% & 82.03\% & 81.33\% & 
82.31\% & 76.55\% & 76.75\% & 74.18\% & 77.86\% & 81.12\%\\
\hline
\multirow{6}{*}{GraD}
& GCN & 86.15\% & 77.72\% &
85.34\% & 85.28\% & 82.32\% & 76.93\% &
80.95\% & 75.30\% & 74.07\% & 68.78\% & 75.41\% & 73.52\%\\
& GAT & 85.13\% & 81.16\% & 
83.89\% & 84.32\% & 82.77\% & 80.19\% &
80.04\% & 72.15\% & 75.04\% & 72.10\% & 69.06\% & 69.40\%\\
& SAGE & 85.59\% & 80.23\% & 
85.87\% & 85.67\% & 84.88\% & 80.97\% & 
79.93\% & 50.40\% & 68.51\% & 59.37\% & 72.40\% & 73.18\%\\
& GNNGuard & 84.67\% & 84.64\% &
84.21\% & 84.02\% & 84.06\% & 84.04\% &
81.41\% & 77.65\% & 78.04\% & 69.72\% & 75.85\% & 76.30\%\\
& RGCN & 85.49\% & 79.51\% & 
84.84\% & 85.05\% & 83.12\% & 78.60\% &
N/A & N/A & N/A & N/A & N/A & N/A\\
& Median & 84.43\% & 82.52\% & 
83.91\% & 84.16\% & 83.85\% & 82.28\% &
81.06\% & 73.85\% & 75.39\% & 71.15\% & 62.99\% & 62.94\%\\
\hline
\label{tab:complete_pubmed_gnns_reduced_acc}
\end{tabular}
\end{table*}

\begin{table*}[ht]
\footnotesize
\centering
\caption{CS's clean accuracy, poisoned accuracy under $p=0.05\%$, and reduced accuracy after graph sparsification under $r=0.325$ with various GNN architecture}
\setlength{\tabcolsep}{2pt} 
\begin{tabular}{cr|rr|rrrr|rrrrrr}
\toprule
\multicolumn{1}{l}{}  & \multicolumn{1}{l|}{} & \multicolumn{1}{l}{} & \multicolumn{1}{l|}{} & \multicolumn{4}{c}{$ACC_r$}\\
\multicolumn{1}{l}{Attack}  & \multicolumn{1}{l|}{GNN} & $ACC_c$ & \multicolumn{1}{l|}{$ACC_p$} & RNE & LD & LS & SCAN & VN & VE & VC & HE & JC & KRON\\
\hline
\multirow{6}{*}{DICE}
& GCN & 92.43\% & 91.81\%
    & 90.91\% & 90.79\% & 91.16\% & 90.68\% & 
    90.12\% & 89.82\% & 90.02\% & 89.89\% & 89.97\% & 89.91\%\\
& GAT & 92.10\% & \textbf{91.56\%}
    & 90.70\% & 90.20\% & 90.94\% & 89.72\% & 
    91.03\% & 91.08\% & 91.15\% & 90.95\% & 91.13\% & 90.86\%\\
& SAGE & 92.13\% & \textbf{91.72\%}
    & \textbf{91.82\%} & 90.50\% & 91.07\% & \textbf{91.92\%} & 
    90.78\% & 90.25\% & 90.31\% & 90.20\% & 89.65\% & 89.56\%\\
& GNNGuard & 92.60\% & \textbf{92.32\%}
    & \textbf{91.90\%} & \textbf{91.94\%} & \textbf{92.12\%} & 91.59\% & 
    91.35\% & 91.17\% & 90.89\% & 90.96\% & 91.38\% & 91.11\%\\    
& RGCN & 92.08\% & \textbf{91.28\%}
    & 90.90\% & 90.75\% & 90.68\% & 90.99\% & 
    N/A & N/A & N/A & N/A & N/A & N/A\\
& Median & 91.91\% & \textbf{91.28\%}
    & 90.80\% & 90.58\% & 90.63\% & \textbf{90.97\%} & 
    90.27\% & 90.31\% & 90.44\% & 90.41\% & 90.34\% & 90.24\%\\
\hline
\multirow{6}{*}{NEA}
& GCN & 92.43\% & \textbf{91.85\%}
    & 90.69\% & 90.80\% & 91.27\% & 90.36\% & 
    90.48\% & 90.18\% & 89.79\% & 90.25\% & 89.81\% & 89.34\%\\
& GAT & 92.10\% & 90.47\%
    & 90.47\% & 90.05\% & 90.97\% & 89.53\% & 
    91.07\% & 91.27\% & 91.04\% & 90.91\% & 91.05\% & 90.93\%\\
& SAGE & 92.13\% & 91.60\%
    & 92.21\% & 91.00\% & 90.74\% & 91.95\% & 
    90.99\% & 90.35\% & 90.02\% & 90.46\% & 89.58\% & 89.35\%\\
& GNNGuard & 92.60\% & \textbf{92.53\%}
    & \textbf{92.23\%} & \textbf{92.10\%} & \textbf{92.27\%} & \textbf{91.61\%} & 
    \textbf{91.74\%} & 91.04\% & 91.24\% & 91.42\% & 91.23\% & 91.37\%\\
& RGCN & 92.08\% & 91.02\%
    & 90.78\% & 90.84\% & 90.93\% & 91.02\% & 
    N/A & N/A & N/A & N/A & N/A & N/A\\
& Median & 91.91\% & \textbf{91.19\%}
    & \textbf{90.97\%} & 90.72\% & 90.68\% & 90.75\% & 
    90.58\% & 90.26\% & 90.01\% & 90.12\% & 90.14\% & 90.08\%\\
\hline
\multirow{6}{*}{PGD}
& GCN & 92.43\% & 88.77\%
    & 88.54\% & 88.37\% & 88.60\% & 88.15\% & 
    87.95\% & 87.29\% & 87.53\% & 87.4\% & 87.11\% & 86.68\%\\
& GAT & 92.10\% & 88.78\%
    & 88.31\% & 88.11\% & 88.56\% & 87.83\% & 
    88.32\% & 88.20\% & 88.03\% & 88.39\% & 88.33\% & 87.89\%\\
& SAGE & 92.13\% & 89.39\%
    & 89.97\% & 88.69\% & 88.56\% & 89.61\% & 
    88.30\% & 88.18\% & 87.73\% & 87.84\% & 87.64\% & 87.4\%\\
& GNNGuard & 92.60\% & 91.51\%
    & 91.16\% & 91.13\% & 91.36\% & 90.90\% & 
    90.83\% & 90.64\% & 90.24\% & 90.5\% & 90.14\% & 90.43\%\\
& RGCN & 92.08\% & 88.33\%
    & 88.08\% & 88.04\% & 88.01\% & 88.14\% & 
    N/A & N/A & N/A & N/A & N/A & N/A\\
& Median & 91.91\% & 88.05\%
    & 88.5\% & 88.05\% & 88.17\% & 88.37\% & 
    88.11\% & 88.02\% & 87.92\% & 87.77\% & 88.04\% & 87.66\%\\
\hline
\multirow{6}{*}{Mettack}
& GCN & 92.43\% & 82.25\%
    & 91.32\% & 90.49\% & 90.53\% & 91.09\% & 
    83.92\% & 84.35\% & 83.24\% & 84.74\% & 83.94\% & 83.73\%\\
& GAT & 92.10\% & 89.26\%
    & 90.79\% & 89.61\% & 90.65\% & 89.79\% & 
    88.50\% & 87.53\% & 87.08\% & 88.87\% & 87.62\% & 88.38\%\\
& SAGE & 92.13\% & 88.10\%
    & 91.82\% & 90.92\% & 90.88\% & 92.21\% & 
    85.32\% & 84.85\% & 84.21\% & 85.69\% & 84.83\% & 85.28\%\\
& GNNGuard & 92.60\% & \textbf{92.25\%}
    & \textbf{92.25\%} & \textbf{91.97\%} & \textbf{92.16\%} & \textbf{92.05\%} & 
    89.04\% & 88.21\% & 86.98\% & 88.83\% & 87.08\% & 88.65\%\\
& RGCN & 92.08\% & 89.02\%
    & 90.56\% & 89.91\% & 89.02\% & 91.01\% & 
    N/A & N/A & N/A & N/A & N/A & N/A\\
& Median & 91.91\% & 90.17\%
    & 90.81\% & 90.32\% & 90.17\% & 91.05\% & 
    85.96\% & 85.72\% & 85.73\% & 85.68\% & 85.87\% & 85.88\%\\
\hline
\multirow{6}{*}{PRBCD}
& GCN & 92.43\% & 88.95\%
    & 88.16\% & 87.77\% & 88.51\% & 87.45\% & 
    87.37\% & 87.50\% & 87.03\% & 87.46\% & 87.39\% & 87.06\%\\
& GAT & 92.10\% & 88.90\%
    & 88.29\% & 87.60\% & 88.59\% & 87.15\% & 
    88.36\% & 88.57\% & 88.40\% & 88.36\% & 88.72\% & 88.58\%\\
& SAGE & 92.13\% & 90.97\%
    & 91.52\% & 89.36\% & 89.78\% & 90.80\% & 
    89.23\% & 89.08\% & 88.76\% & 89.05\% & 88.54\% & 88.54\%\\
& GNNGuard & 92.60\% & \textbf{92.31\%}
    & \textbf{91.90\%} & \textbf{91.74\%} & \textbf{91.94\%} & 91.56\% & 
    90.78\% & 90.72\% & 90.69\% & 91.13\% & 90.72\% & 90.92\%\\
& RGCN & 92.08\% & 88.68\%
    & 88.35\% & 88.41\% & 88.47\% & 88.73\% & 
    N/A & N/A & N/A & N/A & N/A & N/A\\
& Median & 91.91\% & 89.58\%
    & 89.18\% & 88.83\% & 88.92\% & 89.31\% & 
    88.18\% & 88.58\% & 88.21\% & 88.58\% & 88.56\% & 88.50\%\\
\hline
\multirow{6}{*}{STRG-Heuristic}
& GCN & 92.43\% & 91.78\% & 
91.32\% & 91.37\% & 90.89\% & 90.46\% &
88.73\% & 88.73\% & 88.13\% & 87.57\% & 86.63\% & 84.99\%\\
& GAT & 92.10\% & 91.66\% &
91.09\% & 91.14\% & 90.34\% & 89.65\% &
91.01\% & 91.32\% & 90.94\% & 90.90\% & 90.56\% & 90.91\%\\
& SAGE & 92.13\% & 92.21\% &
91.91\% & 91.14\% & 91.07\% & 92.17\% &
88.76\% & 88.65\% & 88.52\% & 87.24\% & 87.26\% & 87.32\%\\
& GNNGuard & 92.60\% & 92.49\% & 
92.06\% & 92.31\% & 92.24\% & 91.75\% & 
91.07\% & 91.26\% & 90.68\% & 90.60\% & 89.56\% & 88.91\%\\
& RGCN & 92.08\% & 89.44\% & 
90.55\% & 90.22\% & 90.10\% & 90.86\% & 
N/A & N/A & N/A & N/A & N/A & N/A\\
& Median & 91.91\% & 90.83\% &
91.04\% & 90.20\% & 90.24\% & 90.71\% &
89.23\% & 89.30\% & 88.55\% & 88.92\% & 88.48\% & 88.46\%\\
\hline
\label{tab:complete_cs_gnns_reduced_acc}
\end{tabular}
\end{table*}

\end{document}